\documentclass[10pt,twocolumn,letterpaper]{article}

\usepackage[pagenumbers]{wacv} %

\usepackage[numbers, sort&compress]{natbib}
\usepackage{graphicx}
\usepackage{amsmath}
\usepackage{amssymb}
\usepackage[utf8]{inputenc} %
\usepackage[T1]{fontenc}    %
\usepackage{url}            %
\usepackage{booktabs}       %
\usepackage{amsfonts}       %
\usepackage{nicefrac}       %
\usepackage{microtype}      %
\usepackage{xcolor}         %
\usepackage{colortbl}
\usepackage{subcaption}
\usepackage{amsfonts}
\usepackage{mathtools}
\usepackage{bbm}
\usepackage{multirow}
\usepackage{wrapfig}
\usepackage{algorithm}
\usepackage{algpseudocode}
\usepackage{algorithmicx}
\usepackage{algcompatible}
\usepackage{xcolor}

\DeclareMathOperator*{\argmin}{argmin} %
\newcommand{\ours}{{\cellcolor[gray]{.9}}}

\usepackage[pagebackref,breaklinks,colorlinks]{hyperref}

\usepackage[capitalize]{cleveref}
\crefname{section}{Sec.}{Secs.}
\Crefname{section}{Section}{Sections}
\Crefname{table}{Table}{Tables}
\crefname{table}{Tab.}{Tabs.}

\def\wacvPaperID{2114} %
\def\confName{WACV}
\def\confYear{2025}

\begin{document}

\title{Hausdorff Distance Matching with Adaptive Query Denoising \\for Rotated Detection Transformer}

\author{Hakjin Lee$^{1\dagger}$, MinKi Song$^{1\dagger}$, Jamyoung Koo$^1$, Junghoon Seo$^{2*}$\\
$^1$SI Analytics Co. Ltd. $^2$KAIST\\
Daejeon, South Korea\\
{\tt\small \{hakjinlee,minkiSong,jmkoo\}@si-analytics.ai, jhseo@kaist.ac.kr}
}
\maketitle

\def\thefootnote{\textdagger}\footnotetext{Equal contribution.}\def\thefootnote{*}\footnotetext{Work performed while at SI Analytics Co. Ltd.}\def\thefootnote{\arabic{footnote}}

\begin{abstract}
Detection Transformers (DETR) have recently set new benchmarks in object detection. However, their performance in detecting rotated objects lags behind established oriented object detectors.
Our analysis identifies a key observation: the boundary discontinuity and square-like problem in bipartite matching poses an issue with assigning appropriate ground truths to predictions, leading to duplicate low-confidence predictions.
To address this, we introduce a Hausdorff distance-based cost for bipartite matching, which more accurately quantifies the discrepancy between predictions and ground truths.
Additionally, we find that a static denoising approach impedes the training of rotated DETR, especially as the quality of the detector's predictions begins to exceed that of the noised ground truths.
To overcome this, we propose an adaptive query denoising method that employs bipartite matching to selectively eliminate noised queries that detract from model improvement.
When compared to models adopting a ResNet-50 backbone, our proposed model yields remarkable improvements, achieving $\textbf{+4.18}$ AP$_{50}$, $\textbf{+4.59}$ AP$_{50}$, and $\textbf{+4.99}$ AP$_{50}$ on DOTA-v2.0, DOTA-v1.5, and DIOR-R, respectively.

\end{abstract}

\section{Introduction}
\label{sec:intro}

In the field of aerial and satellite imagery analysis, the use of oriented bounding boxes is essential for accurately detecting densely packed, arbitrarily-oriented objects with minimal overlap~\cite{xia2018dota, ding2021object}. Various rotated object detectors have been developed by adapting traditional convolution-based object detectors to handle these oriented objects, yielding promising outcomes~\cite{ding2019roitrans,9756223,Ming_Zhou_Miao_Zhang_Li_2021,10.1145/3274895.3274915,xu2023coarsetodynamic,han2021align,han2021redet,xie2021orientedrcnn,yang2021kld,yang2021gwd,yang2023the,Yu_Da_2022,yang2020arbitrary,yang2021dense,Qian_Yang_Peng_Yan_Guo_2021}. This adaptation involves the incorporation of more complex, hand-designed components, such as rotated anchor priors or intricate box representations.

\begin{figure}[tb]
    \centering
    \includegraphics[width=1.0\linewidth]{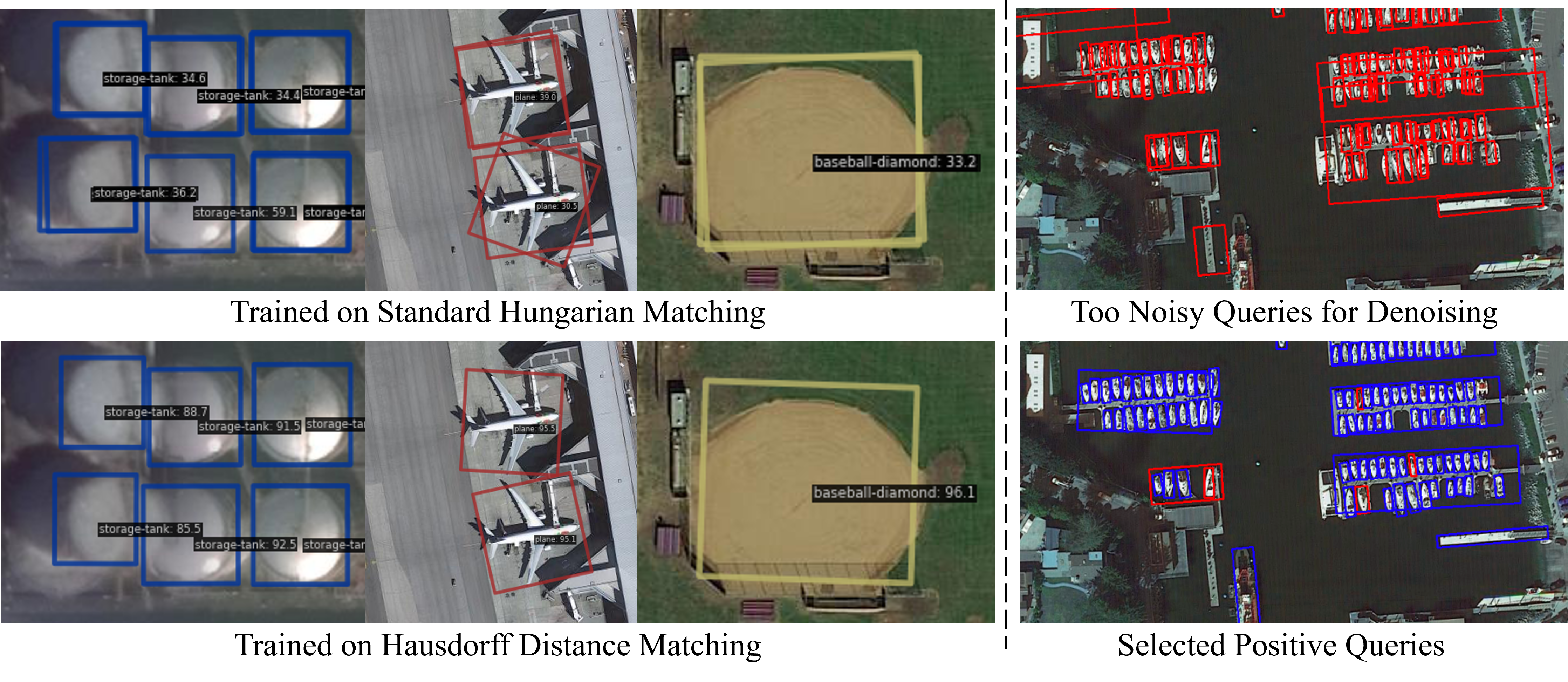}
    \captionof{figure}{Challenges in our rotated detection transformer. \textbf{Left:} The Hausdorff distance addresses the duplicate low-confidence predictions resulting from square-like problem. \textbf{Right:} The proposed adaptive query denoising employs bipartite matching between \textbf{\textcolor{red}{positive noised queries}} and \textbf{\textcolor{blue}{accurate predictions}} to selectively filter queries for denoising.}
\label{fig:main_figure}
\end{figure}

The advent of the DETR has unveiled new possibilities for end-to-end object detection, notably eliminating the need for complex, hand-designed components~\cite{carion2020detr}.
Subsequent advancements have aimed at mitigating DETR's slow training convergence, introducing methods such as deformable attention~\cite{zhu2020deformable}, formulating queries as learnable anchor boxes~\cite{liu2022dab}, and a query denoising task~\cite{liu2022dab,zhang2022dino}.
Notably, DINO~\cite{zhang2022dino} established new benchmarks in the COCO object detection landscape, demonstrating remarkable scalability~\cite{lin2014coco, yang2022focal, zong2022codetr, wang2022internimage}.
Despite these advances, applying DETR to oriented object detection~\cite{ma2021o2d, dai2022ao2, zeng2023ars, zhou2023d2q} reveals that its architecture and performance still do not match those of the most advanced state-of-the-art DETR models.

We speculate that this limited performance arises from challenges in determining the design space of DETR for oriented object detection. For example, while recent DETR models treat their queries as 4D bounding boxes and update them within an output range constrained to [0, 1] using the $\mathrm{sigmoid}$ function, a standard method for integrating the angle variable into this framework remains elusive.
With this challenge in mind, we propose a DINO-like baseline that extends the 4D box representation of reference points and queries to 5D coordinates ($ [x, y, w, h, \theta] $), mapping the output angle from [0, 1] to $ [0^{\circ}, 180^{\circ}] $.
Illustrated in Table~\ref{tab:deformable_baseline}, our frustratingly simple baseline model, without additional modules or novel loss functions, achieves competitive results, rivaling previous oriented detectors based on DETR \cite{ma2021o2d, zeng2023ars, zhou2023d2q, dai2022ao2} as well as convolution-based object detectors \cite{ding2019roitrans,xie2021orientedrcnn}.

Building on this baseline, we identified two critical challenges that hinder the stable training of our baseline DETR model, as depicted in Figure~\ref{fig:main_figure}. The first challenge involves the occurrence of duplicate low-confidence predictions, which contradicts DETRs' essential characteristic of minimizing such duplications. %
In addressing this issue, our analysis of the L1 distance's impact on bipartite matching within rotated DETRs revealed that it leads to spatially inconsistent matching between ground truths and predictions. We propose substituting the L1 distance with the Hausdorff distance, which more accurately quantifies discrepancies between predictions and ground truths. %
The second challenge arises with the query denoising method, particularly when the model's predictions become more accurate than the noised ground truths.
Initially, denoising tasks aid in accelerating DETR training during the early stages. However, in later stages, this becomes problematic as the model persists in learning the distant noised ground truth as supervision, despite making precise predictions.
To resolve this, we introduce an adaptive query denoising strategy that leverages bipartite matching to selectively eliminate these detrimental positive noised queries.
With these advancements, our model, RHINO\footnote{The name RHINO, standing for \textbf{R}otated detection transformer with \textbf{H}ausdorff distance matching, is an homage to D\textbf{INO}.} sets a new state-of-art performance on the DOTA-v1.0/v1.5/v2.0, and DIOR-R benchmarks under single-scale training and testing.

In summary, our research has made the following key contributions:
\begin{enumerate}
\item We design a simple and robust DETR model for rotated object detection, integrating the latest DETR enhancements to deliver competitive performance without the need for additional modules or specialized loss functions.
\item We introduce the Hausdorff distance cost to alleviate the duplicate low-confidence predictions caused by square-like problems in bipartite matching. This method not only resolves duplicate predictions but also leads to more stable training and enhanced performance.
\item We propose an adaptive query denoising method that utilizes bipartite matching to selectively filter out noised ground truths as the model's predictions become more accurate, thereby enhancing performance in the later stages of training.
\end{enumerate}
\section{Related Work}
\label{related_works}
\subsection{Boundary Discontinuity and Square-like Problem in Oriented Object Detection}
Boundary discontinuity refers to the challenge faced when training regression-based object detectors for angle predictions, where an undesirable high loss occurs at the angle range boundary~\cite{9756223, yang2020arbitrary, 9578392, yang2021gwd, cai2024poly, xu2023rethinking, xiao2024theoretically}. For example, a high L1 loss is computed near the angle boundary (e.g. 170$^{\circ}$ prediction versus 10$^{\circ}$ ground truth), despite the bounding boxes having a similar Intersection over Union (IoU).
There are two main approaches to mitigate this: i) transforming angle regression into angle classification and ii) approximating an IoU loss.
CSL~\cite{yang2020arbitrary} addresses this by converting the angle regression challenge into a classification problem with the long edge definition of angles.
DCL~\cite{9578392} enhances the classification-based method by tackling the square-like problem, which induces high losses in objects resembling squares, due to the long edge definition.
In contrast to classification approaches, GWD~\cite{yang2021gwd} and KLD~\cite{yang2021kld} introduce an IoU-like regression loss, substituting the L1 distance-based loss with a distance loss formulated between two 2D Gaussian distributions.
While past research has primarily focused on enhancing regression approaches, our study tackles the challenges associated with suboptimal classification solutions. These challenges emerge when boundary discontinuity and square-like problems meet bipartite matching-based label assignment, a key mechanism for facilitating set prediction in DETR models. Notably, we emphasize the significance of formulating the oriented object detection problem as a 5D parameter regression, especially considering that \textbf{recent DETR enhancements leverage the query-as-bounding-box property by refining queries.}

\begin{figure*}[t!]
    \centering
    \includegraphics[width=0.9\linewidth]{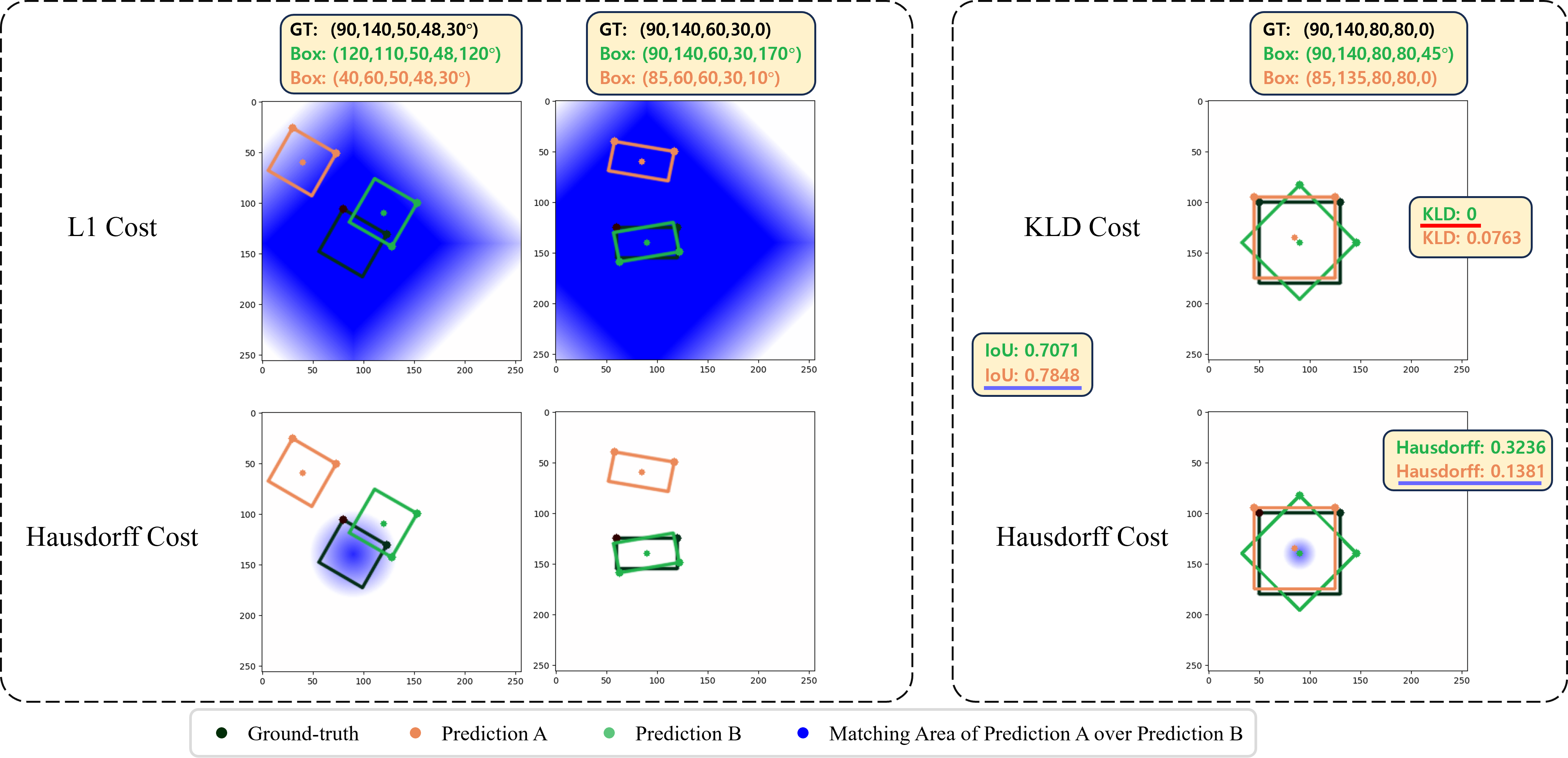}
    \caption{Matching areas of the \textit{Prediction A} to the ground truth. The blue area indicates the orange box is matched to the ground truth over the green box, as the center of the orange box moves along a coordinate axis. In each case, both the ground truth and the green box are fixed. \textbf{Left:} Using L1 cost, the orange box which is too far from the ground truth is matched to it over the green box. \textbf{Right:} When using the KLD cost, the matching ignores the axis alignment between two boxes. Conversely, the Hausdorff cost takes this into account.}
    \label{fig:match_cost}
\end{figure*}

\begin{table}[t!]
    \centering
    \caption{Comparison on DOTA-v1.0 between our baseline and past studies on DETRs for oriented object detection. $^\ast$ indicates multi-scale training and testing.}
    \resizebox{1.0\linewidth}{!}{
            \begin{tabular}[b]{l|c|c|c|cc} 
                \toprule
                Model & Reported by & Backbone & Epochs & AP$_{50}$ & AP$_{75}$ \\ 
                \midrule
                Deformable DETR~\cite{zhu2020deformable} & ~\cite{dai2022ao2} & R-50 & 50 & 41.86 & - \\
                Deformable DETR & ~\cite{zeng2023ars} & R-50 & 50 & 63.42 & 26.92 \\
                \ours Deformable DETR &\ours Ours &\ours R-50 &\ours 50 &\ours 68.50 &\ours 38.13 \\
                \ours Deformable DETR Two-Stage &\ours Ours &\ours R-50 &\ours 50 &\ours 70.48 &\ours 40.07 \\
                \midrule
                O$^2$-DETR$^\ast$ & ~\cite{ma2021o2d} & R-50 & 50 & 72.15 & - \\ 
                ARS-DETR & ~\cite{zeng2023ars} & R-50 & 36 & 73.79 & 49.01 \\
                D2Q-DETR$^\ast$ & ~\cite{zhou2023d2q} & R-50 & 50 & 76.58 & - \\
                AO$^2$-DETR & ~\cite{dai2022ao2} & R-50 & 50 & 77.73 & - \\
                \ours DINO~\cite{zhang2022dino} (Our baseline) &\ours Ours &\ours R-50 &\ours 12 &\ours 76.10 &\ours 48.20 \\
                \ours DINO (Our baseline) &\ours Ours &\ours R-50 & \ours 36 &\ours 74.56 &\ours 49.63 \\
                                \midrule
                Oriented RCNN~\cite{xie2021orientedrcnn} & ~\cite{xie2021orientedrcnn} & R-50 & 12 & 75.87 & -  \\
                RoI Trans.~\cite{ding2019roitrans} & ~\cite{yang2022kfiou} & R-50 & 12 & 75.99 & - \\
                \bottomrule
    \end{tabular}
    }
        \label{tab:deformable_baseline}
\end{table}

\subsection{Query-based Object Detection}
Query-based detectors have recently been extended to tackle rotated object detection tasks in several studies \cite{ma2021o2d,dai2022ao2,zeng2023ars,zhou2023d2q, 10214368dynamiccascade,10387417psdsq,10474035qetr}. AO2-DETR \cite{dai2022ao2} introduces an oriented proposal generation mechanism and an adaptive oriented proposal refinement module, while O2-DETR \cite{ma2021o2d} adopts an efficient encoder for the transformer by replacing the attention mechanism with depthwise separable convolution. ARS-DETR \cite{zeng2023ars} proposes an Aspect Ratio-aware Circle Smooth Label and a rotated deformable attention module, and D2Q-DETR \cite{zhou2023d2q} focuses on decoupling query features and dynamic query design. EMO2-DETR \cite{hu2023emo2} proposes efficient matching to pose a relative redundancy problem of object queries. Although these methodologies have shown that query-based detectors are useful for oriented object detection, the performance of these methods on high-precision oriented object detection tasks still has room for improvement.

\textbf{Query Denoising Strategies.}
The main part of this study is closely related to the query denoising methodology, specifically focusing on DN-DETR \cite{li2022dn} and DINO \cite{zhang2022dino} as two representative works in this area. DN-DETR \cite{li2022dn} introduces a novel denoising training method to accelerate DETR training and address its slow convergence issue. The authors attribute the slow convergence to the instability of bipartite graph matching in the early training stages. By feeding ground-truth bounding boxes with noise into the Transformer decoder and training the model to reconstruct the original boxes, DN-DETR effectively reduces the bipartite graph matching difficulty and achieves faster convergence. DINO \cite{zhang2022dino} builds upon the training method and further enhances performance and efficiency by using a contrastive approach for query denoising. Both of these models have demonstrated significant improvements in the training and performance of DETR-like methods.
In contrast to existing methodologies, our adaptive query denoising approach concentrates on the adaptive reduction of redundant positive queries during the later training phase. This strategy is particularly effective when the model has reached a stage of proficiency where its predictive accuracy surpasses that of the noise-influenced ground truths.

\section{Method}
\label{sec:Approach}

\subsection{Preliminaries}
In this section, we provide a concise overview of set prediction and bipartite matching in the Detection Transformer (DETR). 
In order to enable set prediction without duplicate predictions, the Detection Transformer employs Hungarian loss during its training phase.
In the training phase, a bipartite matching component assigns each prediction result with one ground truth according to the minimum cost between every prediction and ground truth.
The model is then trained to minimize the loss from these assigned results, resulting in predicting a single object for one target, in contrast to other detectors which assign multiple predictions to a single ground truth.
Here, ‘cost’ metrics for bipartite matching and ‘loss’ metrics for training a Detection Transformer are commonly aligned to preserve consistency between the training objective and the strategy for assigning predictions to ground truths.
Several DETR models, including the original DETR, Deformable DETR, DN-DETR, DAB-DETR, and DINO, adopt a combination of metrics for both bipartite matching and the training objective. These metrics include Focal loss~\cite{lin2017focal} for classification, L1 loss, and the Generalized Intersection over Union (GIoU) loss~\cite{rezatofighi2019giou} for box regression. The overall metric is mathematically expressed as:
    \begin{align}
        &\mathcal{L}_\mathrm{train} = \lambda_\mathrm{cls} \cdot \mathcal{L}_\mathrm{cls} + 
        \mathcal{L}_\mathrm{box}, \\
        &\mathcal{L}_\mathrm{box}= \lambda_\mathrm{L1} \cdot \mathcal{L}_\mathrm{L1} + \lambda_\mathrm{iou} \cdot \mathcal{L}_\mathrm{iou},
    \end{align}
where $\lambda$ denotes the weight term assigned to each component to control its impact on the training objective.

\subsection{Rethinking the Baseline for Rotated DETRs}
\label{sec:baseline}
Transitioning from traditional horizontal bounding box detectors to rotated box detectors presents significant challenges.
Recent advancements in DETR, such as iterative reference points refinement~\cite{zhu2020deformable}, the use of queries as dynamic anchor boxes~\cite{liu2022dab}, and the contrastive query denoising training~\cite{zhang2022dino}, require meticulous integration into the rotated DETR frameworks to be effective.

Our primary approach to constructing a generalized DETR baseline for oriented object detection involves the natural incorporation of these components.
We have expanded the standard 4D coordinates $\mathbf{b}=\{b_x, b_y, b_w, b_h\}$ of the regression head output to 5D coordinates $\mathbf{b}=\{b_x, b_y, b_w, b_h, b_\theta\}$, thereby eliminating the need for an additional angle head branch.
The angle values, expressed in radians, are mapped from $ [0, \pi) $ to the range of $[0,1]$, similar to other parameters such as center point coordinates, width, and height.
For the IoU term, $\mathcal{L}_{iou}$, we adopt a KLD loss~\cite{yang2021kld} as the default metric for both the matching cost and IoU loss.
In addition, we have adopted other techniques introduced in Deformable DETR (deformable attention, iterative refinement, and two-stage)~\cite{zhu2020deformable}, DAB-DETR (queries as dynamic anchor boxes)~\cite{liu2022dab}, DN-DETR (denoising training)~\cite{li2022dn}, and DINO (contrastive denoising training, mixed query selection, and look forward twice)~\cite{zhang2022dino}.
Further implementation details can be found in Appendix \ref{sec:supple1_details}.

Table \ref{tab:deformable_baseline} shows the comparison between our implementation and that from the previous works.
Our implementation, even without incorporating domain-specific improvements like rotated feature refinement, exhibits competitive performance. This is noticeable when benchmarked against previous DETR models for oriented object detection on the DOTA-v1.0 dataset~\cite{xia2018dota}.

\begin{figure*}[t!]
    \centering
    \includegraphics[width=0.9\linewidth]{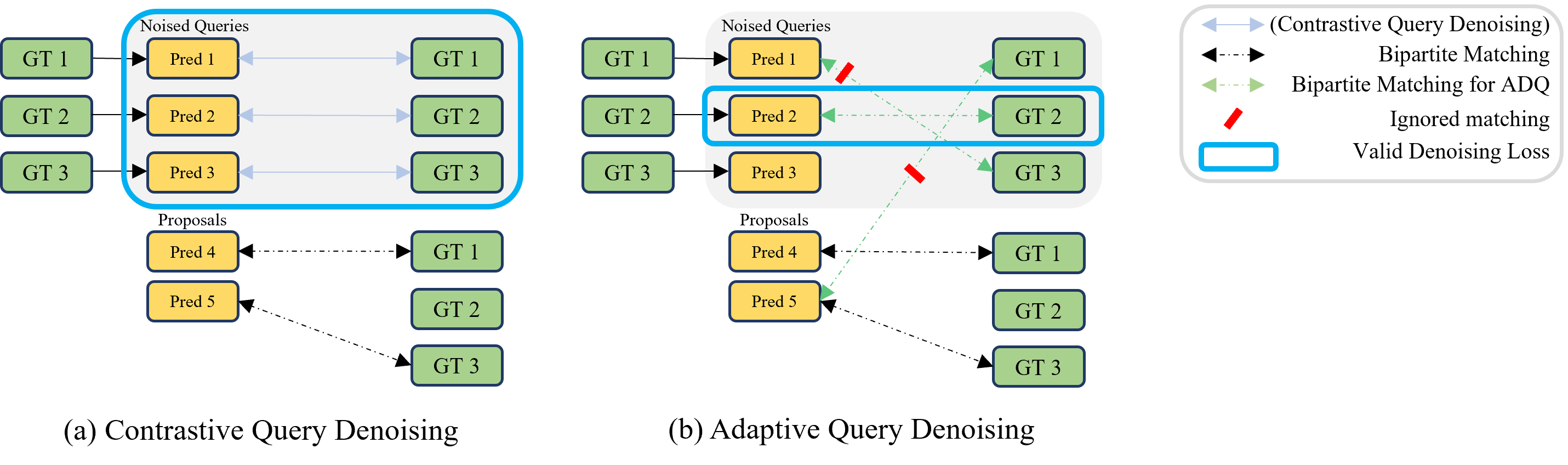}
    \caption{\textbf{Left:} Contrastive query denoising where noised queries and ground truths are directly matched, leading to potential misclassifications. \textbf{Right:} Adaptive query denoising where bipartite matching selectively filters out noised queries, improving the accuracy of predictions as training progresses.}
    \label{fig:adaptive_query_denoising}
\end{figure*}

\subsection{Square-like Problem in Bipartite Matching}
\label{sec:bipartite}

DETR's direct set prediction approach via bipartite matching is well-known to enable non-duplicate predictions without requiring postprocessing such as NMS~\cite{carion2020detr}.
However, our observations indicate a persistent issue of duplicate predictions under specific conditions, as shown in the left image of Figure \ref{fig:main_figure}.
These duplicate predictions frequently occur with lower confidence scores and are particularly noticeable in square-like objects.
The quantitative evidence for this observation is discussed in Table \ref{tab:dup_cost} and Table \ref{tab:aspect} in Appendix \ref{sec:more_able}.

The square-like problem, and related boundary discontinuity problem in oriented object detection, have been demonstrated in previous studies~\cite{yang2021gwd, yang2021dense}.
These studies suggest that the use of common regression loss, such as $l_n$-norms, in training oriented object detectors may lead to an unexpected increase in loss due to angle parameterization.
In the context of bipartite matching in DETR, this issue impacts not only the unexpected regression loss peak but also the classification of the model, as the localization cost plays a crucial role in determining the assignment of predictions to ground truths.

Figure \ref{fig:match_cost} presents common scenarios where the unexpected prediction is matched to the ground truth, despite the presence of a more suitable prediction.
In our analysis, we investigate whether the ground truth is matched to the \textit{Prediction A} (orange box) or the \textit{Prediction B} (green box), under varying conditions.
We compare the matching costs to the ground truth for both the orange box and the green box, where the center of the orange box moves along a coordinate axis.
Before computing the cost, box coordinates $\{b_x, b_y, b_w, b_h, b_\theta\}$ are normalized with the image sizes and the maximum angle value $\{I_w, I_h, I_w, I_h, \pi\}$ for the L1 cost.
The heatmap intensity is highlighted when the cost of matching the orange box to the ground truth is lower than that of matching the green box to the same ground truth.
This indicates that the orange box is preferred for matching the ground truth over the green box when the center of the orange box is within the areas highlighted in blue.
As depicted in Figure \ref{fig:match_cost}, the L1 cost often favors matching the ground truth with the prediction having the same angle, even in instances where the green box exhibits a higher IoU.
This is due to the L1 cost placing excessive emphasis on the angle difference, even after the angle has been normalized.
Consequently, this leads to an unstable matching process.
Predictions that are too far from a certain ground truth, but sharing the same angle, are more likely to have been predicted for the other targets.

\begin{figure}[tb]
    \centering
    \begin{subfigure}{.45\textwidth}
      \centering
      \includegraphics[height=4.2cm]{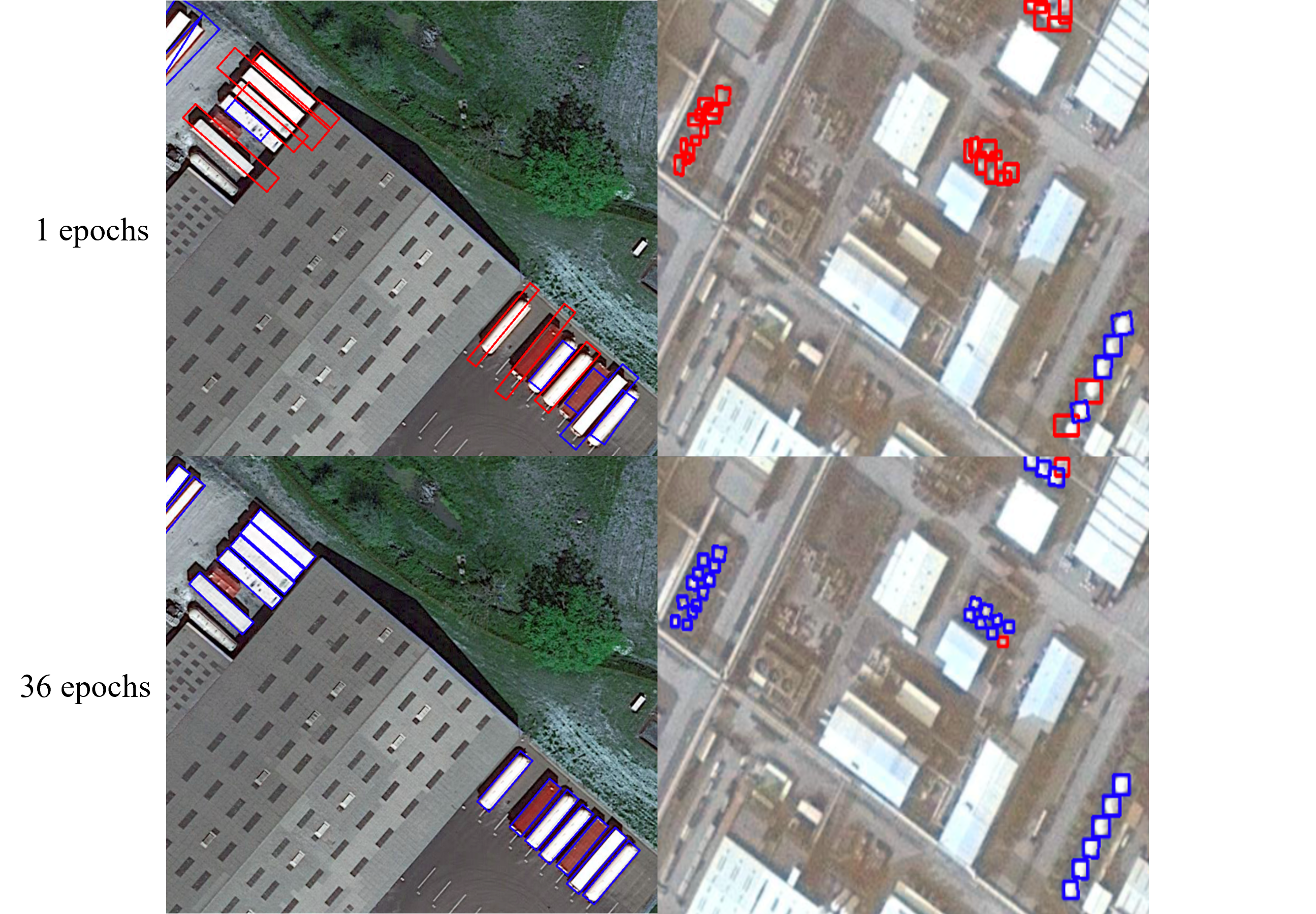}
      \caption{Visualization of used \textcolor{red}{noised queries} for denoising and \textbf{\textcolor{blue}{predictions}}.}
      \label{fig:dynamic_denoising}
    \end{subfigure}
    \hspace{10pt}
    \begin{subfigure}{.45\textwidth}
      \centering
      \includegraphics[height=4.2cm]{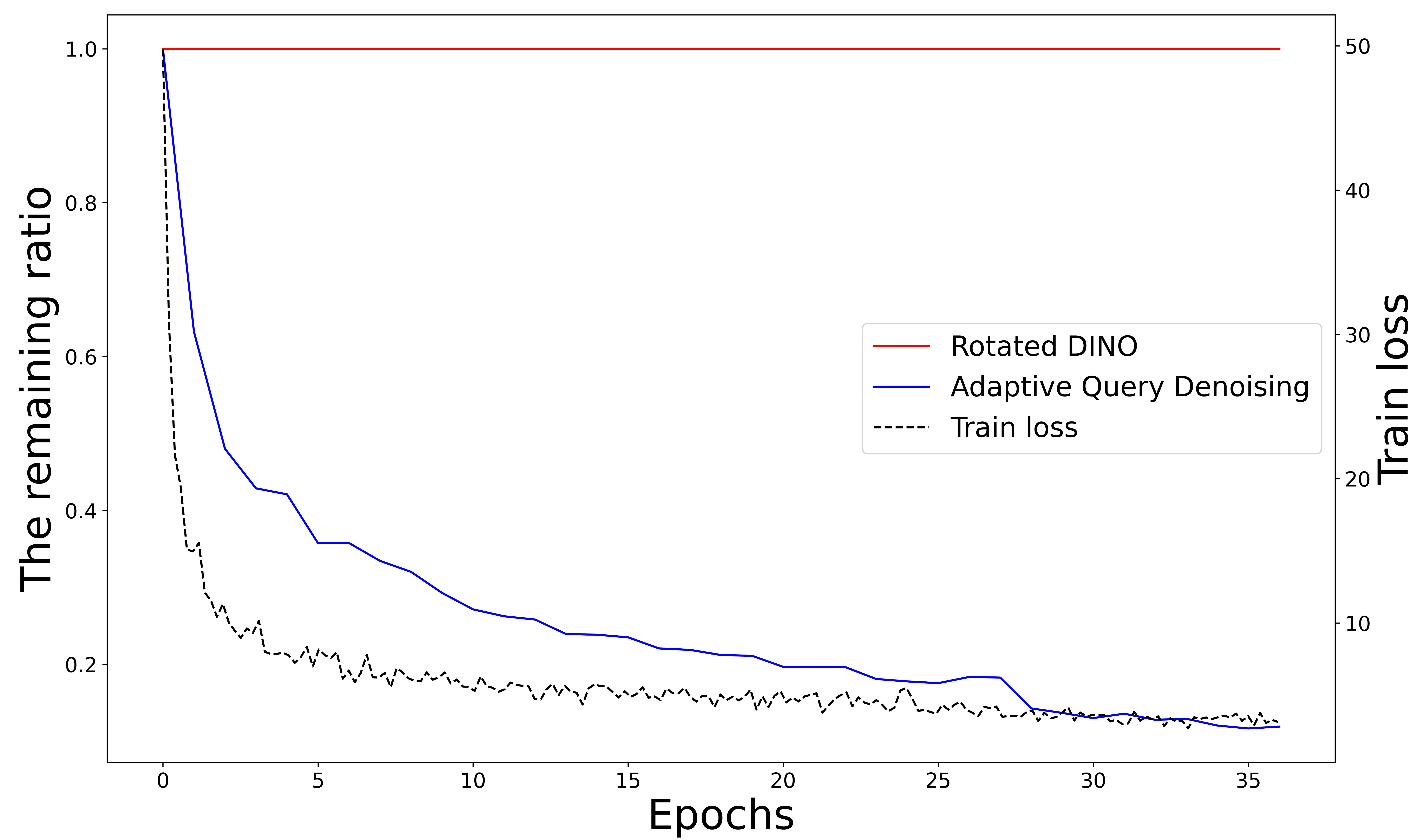}
      \caption{The portion of used noised queries decreases along with the training loss.}
      \label{fig:dn_ratio_plot}
    \end{subfigure}
    \caption{Adaptive query denoising filters out unhelpful noised queries as accuracy improves.}
    \label{fig:adq_full}
\end{figure}

\subsection{Hausdorff Distance for Bipartite Mathcing}
To address the square-like problem and boundary discontinuity caused by the L1 cost, we introduce the Hausdorff distance~\cite{attouch1991topology} for bipartite matching in our model.
The Haussdorff distance is mathematically defined as the maximum distance between two sets of points, which can be expressed as follows:
\begin{equation}
    d_H(X,Y) = \max \biggl\{ \sup_{x \in X} d(x, Y), \sup_{y \in Y} d(X, y) \biggr\},
\end{equation}
where $d(x, Y) = \inf_{y \in Y} d(x, y)$ denotes the distance from a point $x$ to the point set $Y$, and $d(X, y) = \inf_{x \in X} d(x, y)$ represents the distance from a point $y$ to the point set $X$. The function $d(x,y)$ measures the distance between two points $x$ and $y$, which is commonly defined as the Euclidean distance in our implementation. In the simplest form, we can represent the bounding box by the set of its four corner coordinates. Using this representation, it is possible to calculate the maximum value of pairwise distances between two boxes.
In a more general form, we can determine the midpoints of each edge of the bounding box and use these midpoints to calculate the pairwise maximum distance.
For the Hausdorff distance, box coordinates are normalized using the factors  $\{I_w, I_h, I_w, I_h, 1\}$ to preserve the angle values.
As shown in Figure \ref{fig:match_cost}, using the Hausdorff cost leads to more reasonable matching compared to the L1 cost.
Moreover, the Hausdorff distance considers the axis alignment of two boxes, resulting in a more accurate matching compared to 2D Gaussian distribution-based metrics such as KLD~\cite{yang2021kld} and GWD~\cite{yang2021gwd}.

\begin{table*}[t]
\caption{Main results on DOTA-v2.0 test set under single-scale training and testing. \textbf{\color{red}{Red}} and \textbf{\color{blue}{blue}}: top two performances.}
\label{tab:dotav2}
\centering
\resizebox{1\textwidth}{!}{
\begin{tabular}{l|c|cccccccccccccccccc|c}  
	\toprule
	Method  & Backbone & PL & BD & BR & GTF & SV & LV & SH & TC & BC & ST & SBF & RA & HA & SP & HC & CC & AP & HL & AP$_{50}$  \\
    \midrule
    $\rm{R^3Det}$~w/~KLD~\cite{yang2021kld} & R-50 & - & - & - & - & - & - & - & - & - & - & - & - & - & - & - & - & - & - & 50.90 \\
    Roi Trans.~\cite{ding2019roitrans} & R-50 &  71.81 & 48.39 & 45.88 & 64.02 & 42.09 & 54.39 & 59.92 & 82.70 & 63.29 & 58.71 & 41.04 & 52.82 & 53.32 & 56.18 & 57.94 & 25.71 & 63.72 & 8.70 & 52.81 \\
    
    Oriented R-CNN~\cite{xie2021orientedrcnn} & R-50 & 77.95 & 50.29 & 46.73 & 65.24 & 42.61 & 54.56 & 60.02 & 79.08 & 61.69 & 59.42 & 42.26 & 56.89 & 51.11 & 56.16 & 59.33 & 25.81 & 60.67 & 9.17 & 53.28\\

    DCFL~\cite{xu2023coarsetodynamic}   & R-50 & 78.30 & 53.03 & 44.24 & 60.17 & 48.56 & 55.42 & 58.66 & 78.29 & 60.89 & 65.93 & 43.54 & 55.82 & 53.33 & 60.00 & 54.76 & 30.90 & 74.01 & 15.60 & 55.08 \\
    DCFL~\cite{xu2023coarsetodynamic}   & ReR-101 & 79.49 & 55.97 & 50.15 & 61.59 & 49.00 & 55.33 & 59.31 & 81.18 & 66.52 & 60.06 & 52.87 & 56.71 & 57.83 & 58.13 & 60.35 & 35.66 & 78.65 & 13.03 & 57.66 \\
    RoI Trans. w/ ~\cite{cha2023billion}  & ViT-G12$\times$4 (MAE) & 80.12 & 54.12 & 50.07 & 65.68 & 43.98 & 60.07 & 67.85 & 79.11 & 64.38 & 60.56 & 45.98 & 58.26 & 58.31 & 64.82 & 69.84 & 32.78 & 89.37 & 11.07 & 58.69 \\
    \midrule
        \ours RHINO (Ours) & \ours R-50 & \ours 79.32 & \ours 58.33 & \ours 46.06 & \ours 65.79 & \ours 57.25 & \ours 58.38 & \ours 68.53 & \ours 80.46 & \ours 63.27 & \ours 73.01 & \ours 48.69 & \ours 57.80 & \ours 56.21 & \ours 60.04 & \ours 59.06 & \ours 42.70 & \ours 78.96 & \ours 12.84 & \ours \textbf{\textcolor{blue}{59.26}} \\
        \ours RHINO (Ours) & \ours Swin-T & \ours 79.74 & \ours 58.79 & \ours 48.13 & \ours 67.12 & \ours 57.21 & \ours 59.11 & \ours 69.48 & \ours 83.54 & \ours 65.14 & \ours 74.05 & \ours 47.93 & \ours 60.49 & \ours 58.43 & \ours 63.25 & \ours 55.59 & \ours 48.49 & \ours 82.06 & \ours 14.40 & \ours \textbf{\textcolor{red}{60.72}} \\
	\bottomrule
	\end{tabular}}
\end{table*}
\begin{table*}[t]
\caption{Main results on DOTA-v1.5 test set under single-scale training and testing. \textbf{\color{red}{Red}} and \textbf{\color{blue}{blue}}: top two performances.}
\label{tab:dotav15}
        \centering
		\resizebox{1.0\textwidth}{!}{
    \begin{tabular}{l|c|cccccccccccccccc|c} 
    \toprule
    Method & Backbone & PL & BD & BR & GTF & SV & LV & SH & TC & BC & ST & SBF & RA & HA & SP & HC & CC & AP$_{50}$ \\ \midrule
    EMO2-DETR~\cite{hu2023emo2}     &R-50&71.81&75.36&45.09&58.70&48.19&73.26&80.28&90.70&73.05&76.53&39.36&65.31&56.96&69.29&47.11&15.64&61.67\\
    Mask R-CNN~\cite{he2017mask}     &R-50&76.84&73.51&49.90&57.80&51.31&71.34&79.75&90.46&74.21&66.07&46.21&70.61&63.07&64.46&57.81&9.42&62.67\\
    HTC~\cite{chen2019hybrid}            &R-50&77.80&73.67&51.40&63.99&51.54&73.31&80.31&90.48&75.12&67.34&48.51&70.63&64.84&64.48&55.87&5.15&63.40\\
    AO$^2$-DETR~\cite{dai2022ao2} & R-50 & 79.55 & 78.14 & 42.41 & 61.23 & 55.34 & 74.50 & 79.57 & 90.64 & 74.76 & 77.58 & 53.56 & 66.91 & 58.56 & 73.11 & 69.64 & 24.71 & 66.26 \\
ReDet~\cite{han2021redet}&ReR-50&79.20&82.81&51.92&71.41&52.38&75.73&80.92&90.83&75.81&68.64&49.29&72.03&73.36&70.55&63.33&11.53&66.86 \\
    DCFL~\cite{xu2023coarsetodynamic} &R-50&-&-&-&-&56.72&-&80.87&-&-&75.65&-&-&-&-&-&-&67.37 \\
    DCFL~\cite{xu2023coarsetodynamic} &ReR-101&-&-&-&-&52.38&-&86.60&-&-&76.55&-&-&-&-&-&-& 70.24 \\
    Oriented R-CNN w/~\cite{cai2024poly} &PKINet-S&80.31&85.00&55.61&74.38&52.41&76.85&88.38&90.87&79.04&68.78&67.47&72.45&76.24&74.53&64.07&37.13& 71.47 \\
    \midrule
    \ours RHINO (Ours) & \ours R-50 & \ours 77.96&\ours 83.22&\ours 55.30&\ours 72.14&\ours 65.07&\ours 78.95&\ours 89.22&\ours 90.78&\ours 80.90&\ours 83.48&\ours 61.58&\ours 74.17&\ours 77.21&\ours 71.99&\ours 60.16&\ours 29.26& \ours \textbf{\textcolor{blue}{71.96}} \\
    \ours RHINO (Ours) & \ours Swin-T & \ours 79.96 & \ours 85.16 & \ours 56.96 & \ours 74.50 & \ours 64.14 & \ours 81.91 & \ours 89.42 & \ours 90.60 & \ours 83.93 & \ours 82.89 & \ours 59.93 & \ours 74.84 & \ours 77.23 & \ours 75.18 & \ours 64.91 & \ours 33.76 & \ours \textbf{\textcolor{red}{73.46}}
 \\\bottomrule
    \end{tabular}%
    }
    \vspace{-2mm}
\end{table*}

\begin{table*}[t]
  \setlength{\belowcaptionskip}{-6pt}
    \caption{Main results on DOTA-v1.0 test set under single-scale training and testing. ReR-50 indicates ReResNet-50~\cite{han2021redet}. %
    }
    \label{tab:dotav1}
    \begin{center}
        \resizebox{1.0\textwidth}{!}{
            \begin{tabular}{l|c|c|c|c|c|c|c|c|c|c|c|c|c|c|c|c|c} 
            \toprule
            Method & Backbone & PL & BD & BR & GTF & SV & LV & SH & TC & BC & ST & SBF & RA & HA & SP & HC & AP$_{50}$  \\ 
            \midrule

            EMO2-DETR~\cite{hu2023emo2} & R-50 & 88.08 & 77.91 & 43.17 & 62.91 & 74.01 & 75.09 & 79.21 & 90.88 & 81.50 & 84.04 & 51.92 & 59.44 & 64.74 & 71.81 & 58.96 & 70.91 \\
            EMO2-DETR~\cite{hu2023emo2} & Swin-T & 89.03 & 79.59 & 48.71 & 60.23 & 77.34 & 76.42 & 84.53 & 90.77 & 84.80 & 85.68 & 48.86 & 67.55 & 66.32 & 71.54 & 53.49 & 72.32  \\
            ARS-DETR~\cite{zeng2023ars} & R-50 & 86.61 & 77.26 & 48.84 & 66.76 & 78.38 & 78.96 & 87.40 & 90.61 & 82.76 & 82.19 & 54.02 & 62.61 & 72.64 & 72.80 & 64.96 & 73.79 \\
            ARS-DETR~\cite{zeng2023ars} & Swin-T & 87.78 & 78.58 & 52.58 & 67.69 & 80.19 & 84.32 & 88.19 & 90.68 & 85.92 & 84.76 & 55.18 & 66.89 & 74.57 & 79.09 & 60.35 & 75.79  \\
            RoI Trans.~\cite{ding2019roitrans} (reported by ~\cite{yang2022kfiou}) & R-50 & 89.02 & 81.71 & 53.84 & 71.65 & 79.00 & 77.76 & 87.85 & 90.90 & 87.04 & 85.70 & 61.73 & 64.55 & 75.06 & 71.71 & 62.38 & 75.99 \\

            ReDet~\cite{han2021redet} & ReR-50 & 88.79 & 82.64 & 53.97 & 74.00 & 78.13 & 84.06 & 88.04 & 90.89 & 87.78 & 85.75 & 61.76 & 60.39 & 75.96 & 68.07 & 63.59 & 76.25 \\

            Oriented RCNN~\cite{xie2021orientedrcnn} & R-50 & 89.46 & 82.12 & 54.78 & 70.86 & 78.93 & 83.00 & 88.20 & 90.90 & 87.50 & 84.68 & 63.97 & 67.69 & 74.94 & 68.84 & 52.28 & 75.87  \\
            Oriented RCNN~\cite{xie2021orientedrcnn} & R-101 & 88.86 & 83.48 & 55.27 & 76.92 & 74.27 & 82.10 & 87.52 & 90.90 & 85.56 & 85.33 & 65.51 & 66.82 & 74.36 & 70.15 & 57.28 & 76.28  \\
            
            R$^3$Det-KLD~\cite{yang2021kld} & R-50 & 88.90 & 84.17 & 55.80 & 69.35 & 78.72 & 84.08 & 87.00 & 89.75 & 84.32 & 85.73 & 64.74 & 61.80 & 76.62 & 78.49 & 70.89 & 77.36 \\
            
            AO$^2$-DETR~\cite{dai2022ao2} & R-50 & 89.27 & 84.97 & 56.67 & 74.89 & 78.87 & 82.73 & 87.35 & 90.50 & 84.68 & 85.41 & 61.97 & 69.96 & 74.68 & 72.39 & 71.62 & 77.73  \\
            RoI Trans. w/ KFIoU~\cite{yang2022kfiou} & Swin-T & 88.9 & 83.77 & 53.98 & 77.63 & 78.83 & 84.22 & 88.15 & 90.91 & 87.21 & 86.14 & 67.79 & 65.73 & 75.80 & 73.68 & 63.30 & 77.74  \\
            RTMDet-R-m~\cite{lyu2022rtmdet} & CSPNeXt-m & 89.17 & 84.65 & 53.92 & 74.67 & 81.48 & 83.99 & 88.71 & 90.85 & 87.43 & 87.20 & 59.39 & 66.68 & 77.71 & 82.40 & 65.28 & 78.24  \\
            RTMDet-R-l~\cite{lyu2022rtmdet} & CSPNeXt-l & 89.43 & 84.21 & 55.20 & 75.06 & 80.81 & 84.53 & 88.97 & 90.90 & 87.38 & 87.25 & 63.09 & 67.87 & 78.09 & 80.78 & 69.13 & 78.85  \\
            PSD-SQ~\cite{10387417psdsq} & R-50 & 89.67 & 85.43 & 57.28 & 75.24 & 79.98 & 81.21 & 88.30 & 89.88 & 88.05 & 86.35 & 69.63 & 68.45 & 75.33 & 69.46 & 72.65 & 78.46  \\
            Oriented R-CNN w/~\cite{wang2022vitaers} & ViTAE-B + RVSA(MAE) & 89.38 & 84.26 & 59.39 & 73.19 & 79.99 & 85.36 & 88.08 & 90.87 & 88.50 & 86.53 & 58.93 & 72.24 & 77.31 & 79.59 & 71.24 & \textbf{\textcolor{blue}{78.99}}  \\
            \midrule
            \ours RHINO (Ours) & \ours R-50 & \ours 88.28 & \ours 85.19 & \ours 55.89 & \ours 72.75 & \ours 80.23 & \ours 83.14 & \ours 89.07 &\ours 90.88 & \ours 87.12 & \ours 86.89 & \ours 65.37 & \ours 71.68 & \ours 77.75 & \ours 81.24 & \ours 64.71 & \ours 78.68 \\
            \ours RHINO (Ours) & \ours Swin-T & \ours 88.18 & \ours 84.77 & \ours 58.54 & \ours 77.73 & \ours 81.13 & \ours 85.58 & \ours 89.18 & \ours 90.86 & \ours 86.96 & \ours 86.38 & \ours 65.49 & \ours 71.29 & \ours 78.17 & \ours 82.80 & \ours 64.26 & \ours \textbf{\textcolor{red}{79.42}} \\
            \bottomrule
            \end{tabular}}
        \end{center}
        \vspace{-6mm}
\end{table*}

\subsection{Adaptive Denoising via Bipartite Matching}
\label{sec:Adaptive Query Denoising via Bipartite Mathcing}
Query denoising training~\cite{li2022dn} introduces a refinement task that aims to refine noised queries generated from the ground truth. The loss is computed between these noised queries and their corresponding ground truth.
We visualize the noised queries and the query proposals produced by a model trained for 36 epochs in the right panel of Figure~\ref{fig:main_figure}.
It is clear that a significant portion of positive noised queries are not ideally suited for accurate classification.
For instance, in the right panel of Figure~\ref{fig:main_figure}, a positive query located outside the ship should be classified as a ship.
To filter out unhelpful positive noised queries as training progresses, we propose an adaptive query denoising using bipartite matching.
Figure \ref{fig:adaptive_query_denoising} illustrates a comparison between contrastive query denoising~\cite{zhang2022dino} and our proposed adaptive query denoising.
After matching ground truths to both noised queries and proposals, only the noised queries that match the source ground truth are used for the denoising loss.

We denote predictions as $\mathbf{p}=\{p_0, p_1, \dots, p_{N-1} \}$, where $N$ is the number of predicted objects, and ground truths as $\mathbf{y}=\{y_0, y_1, \dots, y_{M-1}\}$ for $M$ ground truth objects.
We refer to noised queries in a denoising group as $\mathbf{q}^\mathrm{pos}=\{q^\mathrm{pos}_0, q^\mathrm{pos}_1, \dots, q^\mathrm{pos}_{M-1} \}$ for positive queries and $\mathbf{q}^\mathrm{neg}=\{q^\mathrm{neg}_0, q^\mathrm{neg}_1, \dots, q^\mathrm{neg}_{M-1} \}$ as negative queries, which are generated from $M$ ground truths by adding noise.
Not to filter out accurately refined queries, we use refined noised queries through decoder layers $\mathbf{p}^\mathrm{pos}=f_\mathrm{refine}(\mathbf{q}^\mathrm{pos})$ and  $\mathbf{p}^\mathrm{neg}=f_\mathrm{refine}(\mathbf{q}^\mathrm{neg})$ instead of directly using $\mathbf{q}^\mathrm{pos}$ and $\mathbf{q}^\mathrm{neg}$.
Each element $p_i,y_j, p^\mathrm{pos}_k, p^\mathrm{neg}_l $ is composed of $(c, \mathbf{b})$ where $c$ is the target class and $\mathbf{b}=\{b_x, b_y, b_w, b_h, b_\theta\}$ is a 5D rotated box coordinate.
The reconstruction loss in a single denoising group can be formulated as follows:
\begin{equation}
\begin{aligned}
\mathcal{L}_\mathrm{denoising}(\mathbf{p}^\mathrm{pos}, \mathbf{p}^\mathrm{neg}, \mathbf{y})
&= \sum_{i=0}^{M-1} \mathcal{L}_\mathrm{pos}(p^\mathrm{pos}_{i}, y_i) \\
&\quad + \mathcal{L}_\mathrm{neg}(p_i^\mathrm{neg}, \varnothing),
\end{aligned}
\end{equation}
where $\mathcal{L}_\mathrm{neg}(p_i^\mathrm{neg}, \varnothing)$ indicates a classification loss such as Focal loss for no object $\varnothing$.

To remove excessively noisy positive queries, $\mathcal{L}_\mathrm{pos}(p^\mathrm{pos}_{i}, y_i)$ is partitioned into two parts: the base training objective $\mathcal{L}_\mathrm{train}$ and the background loss $\mathcal{L}_\mathrm{neg}$ for filtered queries as below:

\begin{equation}
\label{eq:dd}
\begin{aligned}
\mathcal{L}_\mathrm{pos}(p^\mathrm{pos}_{i}, y_i)
&= \mathbbm{1}_{\{\hat{\sigma}_i = i\}}\mathcal{L}_\mathrm{train}(p^\mathrm{pos}_{i}, y_i) \\
&\quad + \mathbbm{1}_{\{\hat{\sigma}_i \neq i\}}\mathcal{L}_\mathrm{neg}(p^\mathrm{pos}_{i}, \varnothing),
\end{aligned}
\end{equation}
where $\hat{\sigma}_i$ indicates the matching index for a ground truth $y_i$.
The matching indices are determined by finding the optimal assignment between ground truths and a concatenated list of positive queries and predictions:

\begin{equation}
\begin{aligned}
\mathbf{p}^\mathrm{all} &= [\mathbf{p}^\mathrm{pos}; \mathbf{p}] \\
    &= [p^\mathrm{pos}_0, p^\mathrm{pos}_1, \dots, p^\mathrm{pos}_{M-1}, \\
    &\quad\quad p_0, p_1, \dots, p_{N-1}],
\end{aligned}
\end{equation}

\begin{gather}
    \hat{\sigma} =\argmin_{\sigma \in \mathfrak{S}}\sum_{i=0}^{M+N-1} \mathcal{L}_\mathrm{match}(\hat{y}_i, \mathbf{p}^\mathrm{all}_{\sigma_i}),
\end{gather}
where $\mathfrak{S}$ is the set of all permutations of size $M+N-1$ and $\hat{y}_i$ is a padded version of ground truth objects. 
$\mathcal{L}_\mathrm{match}(\hat{y}_i, \mathbf{p}^\mathrm{all}_{\sigma_i})$ represents a pair-wise matching cost between $\hat{y}_i$ and $\mathbf{p}^\mathrm{all}$ with index $\sigma_i$.
This indicates that only positive queries whose matching index is the same as the index where they were generated from the ground truth are used in the base training objective.
As illustrated in Figure~\ref{fig:dynamic_denoising} and Figure~\ref{fig:dn_ratio_plot}, the proposed adaptive query denoising effectively filters out unhelpful noised queries as the predictions become more accurate through training.

Since our rationale for filtering out unhelpful queries is based on the classification task, we developed an alternative method, named \textit{improved adaptive query denoising}, which maintains a regression loss for the filtered positive queries. This approach is formulated as:
\begin{equation}
\begin{aligned}
    \label{eq:improved_dd}
    \mathcal{L}^*_\mathrm{pos}(p^\mathrm{pos}_{i}, y_i) 
    &= \mathcal{L}_\mathrm{pos}(p^\mathrm{pos}_{i}, y_i) \\
    &\quad + \mathbbm{1}_{\{\hat{\sigma}_i \neq i\}} \mathcal{L}_\mathrm{bbox}(p^\mathrm{pos}_{i}, y_i).
\end{aligned}
\end{equation}
Table \ref{tab:components_ablation} demonstrated maintaining a regression loss for the filtered positive queries could be still beneficial to the model.
This strategy helps guide the model in refining its localization of predictions, especially proposals that are initially less accurate. Algorithm ~\ref{algo:aqd} in Appendix illustrates the pseudocode of our adaptive query denoising method.
\section{Experiments}\label{sec:Experiments}

\subsection{Experiment Setting}
\textbf{Dataset.} We evaluate our proposed models on four datasets: DOTA-v1.0~\cite{xia2018dota} / v1.5 / v2.0~\cite{ding2021object}, and DIOR-R~\cite{cheng2022diorr} under single-scale training and testing setting.
DOTA-v1.0 / v1.5 / v2.0 differ in the number of categories and the size of the training and test dataset.
Unless otherwise specified, ablation studies are conducted on DOTA-v1.0.
For all datasets, we use {\tt train set, val set} for training and {\tt test set} for evaluation. Evaluations on DOTA-v1.0/v1.5/v2.0 are conducted in the official DOTA evaluation server.

\textbf{Implementation Details.} The implementation details of our final model, RHINO, are described in Appendix \ref{sec:imple_rhino}. Unless explicitly stated otherwise, we adhere to the default hyperparameters as outlined in DINO~\cite{zhang2022dino}. In our default configuration, the Hausdorff is computed using the four corner points of the bounding box coordinates.

\subsection{Main Results}
\label{sec:exp-main}
We compare our models with their counterparts that use an ImageNet pre-trained ResNet-50 backbone on DOTA-v2.0, DOTA-v1.5, DOTA-v1.0, and DIOR-R in order.
Table \ref{tab:dotav2} presents the performance of our RHINO model against state-of-the-art models on DOTA-v2.0, which is one of the most comprehensive datasets in oriented object detection. 
RHINO with a ResNet-50 backbone outperforms all other counterparts by achieving a \textbf{+4.18} AP$_{50}$ improvement, surpassing even those models with advanced backbones like ReResNet~\cite{han2021redet} or ViT-G12$\times$4 (MAE)\cite{dosovitskiy2020vit}. Furthermore, RHINO with a Swin-Tiny sets a new benchmark with \textbf{60.72} AP$_{50}$.
Table \ref{tab:dotav15} shows a comparison of RHINO against state-of-the-art models on the DOTA-v1.5 dataset. Among models with a ResNet-50 backbone, RHINO stands out with a significant performance improvement of \textbf{+4.59} AP$_{50}$.  Utilizing a Swin-T backbone further extends its lead, reaching \textbf{73.46} AP$_{50}$ and surpassing all other models.
On the DOTA-v1.0 dataset, Table \ref{tab:dotav1} presents a comparison under single-scale training and testing, showcasing RHINO's impressive performance. Among models equipped with a ResNet-50 backbone, RHINO reaches \textbf{78.68} AP$_{50}$, marking a \textbf{+0.96} AP$_{50}$ gap above its closest competitors. With a Swin-T backbone, RHINO further achieves a higher score of 79.42 AP$_{50}$.
For DIOR-R, Table \ref{tab:dior_r} shows RHINO with ResNet-50 surpasses other models with a ResNet-50 and even the model with a heavier and domain-specific pre-trained backbone (ReR-101 and ViTAE-B + RVSA (MAE)~\cite{wang2022vitaers}).
Furthermore, using Swin-Tiny as a backbone leverages the performance to \textbf{72.67} AP$_{50}$. The speed and the number of parameters of our model are reported in Appendix ~\ref{sec:cost-speed}.

\begin{table}[t]
    \caption{Performance comparison on DIOR-R. %
    }
    \label{tab:dior_r}
        \centering
        \scalebox{.65}{
            \begin{tabular}{l|c|c} 
                \toprule
                Method & Backbone & AP$_{50}$ \\ 
                \midrule
                Rotated RetinaNet~\cite{lin2017focal} & R-50 & 57.55 \\
                Rotated Faster R-CNN~\cite{ren2015faster} & R-50 & 59.54 \\
                RoI Trans.~\cite{ding2019roitrans} & R-50 & 63.87 \\
                DCFL~\cite{xu2023coarsetodynamic} & R-50 & 66.80 \\
                DCFL~\cite{xu2023coarsetodynamic} & ReR-101 & 71.03 \\
                Oriented R-CNN w/~\cite{cai2024poly} & PKINet-S & 67.03 \\
                Oriented R-CNN w/~\cite{wang2022vitaers} & ViTAE-B + RVSA(MAE) & 71.05 \\
                \midrule
                \ours RHINO (Ours) & \ours R-50 & \ours \textbf{\textcolor{blue}{71.79}}  \\
                \ours RHINO (Ours) & \ours Swin-T & \ours \textbf{\textcolor{red}{72.67}}  \\
                \bottomrule
            \end{tabular}}
\end{table}

\begin{table}[t]
    \caption{High-precision comparison on DOTA-v1.0.} %
    \label{tab:dota1_ap75}
        \centering
        \scalebox{.65}{
            \begin{tabular}{l|c|c|c|c} 
                \toprule
                Method & Backbone & \# points & AP$_{50}$ & AP$_{75}$ \\ 
                \midrule
                Oriented RCNN~\cite{xie2021orientedrcnn} & R-50 & - &  74.19 & 46.96 \\

                RoI Trans.~\cite{ding2019roitrans} & R-50 & - & 74.05 & 46.54 \\
                RoI Trans.~\cite{ding2019roitrans} & Swin-T & - & 76.49 & 50.15  \\
                ReDet~\cite{han2021redet} & ReR-50 & - & 76.25 & 50.86 \\
                ARS-DETR~\cite{zeng2023ars} & R-50 & - & 73.79 & 49.01 \\
                ARS-DETR~\cite{zeng2023ars} & Swin-T & - & 75.79 & 51.11 \\
                \midrule
                \ours RHINO w/ KLD & \ours R-50 & \ours32 & \ours \textbf{\textcolor{red}{78.49}} & \ours 51.84 \\
                \ours RHINO w/ GWD & \ours R-50 & \ours32 & \ours \textbf{\textcolor{blue}{77.87}} & \ours 51.49 \\
                \ours RHINO w/ RIoU & \ours R-50 & \ours32 & \ours 77.68 & \ours \textbf{\textcolor{blue}{53.51}} \\
                \ours RHINO w/ GRIoU & \ours R-50 & \ours32 & \ours 77.24 & \ours \textbf{\textcolor{red}{53.91}} \\
                \bottomrule
            \end{tabular}}
\end{table}

\begin{table*}[t]
\caption{Ablation studies of model architecture on DOTA-v1.0.}
\label{tab:ablation_all}
    \centering
    \begin{minipage}[t]{.4\linewidth}
            \subcaption{Effect of each proposed component}
        \label{tab:components_ablation}
        \centering
            \scalebox{0.8}{\begin{tabular}{cccc|cc} 
                \toprule
                DINO & Hausdorff & AQD & AQD* & AP$_{50}$  & AP$_{75}$ \\ 
                \midrule
                ~ & ~ & ~ & ~& 70.48 & 40.78 \\
                \checkmark & ~ & ~ & ~ & 74.56 & 49.63  \\
                \checkmark  & \checkmark & ~ & ~  & 76.14  & 50.33  \\
                \checkmark & ~ & \checkmark & ~  & 75.93 & 49.84 \\
                \checkmark & \checkmark & \checkmark & ~  & 78.05 & 50.59 \\
                \checkmark & ~ & ~ & \checkmark & 76.63  & 50.97 \\
                \checkmark & \checkmark & ~ & \checkmark  & \textbf{78.68} & \textbf{51.17} \\
                \bottomrule
            \end{tabular}}
    \end{minipage}
    \begin{minipage}[t]{.35\linewidth}
            \subcaption{Duplicate prediction comparison with $\mathcal{L}_\mathrm{L1}$.}
        \label{tab:dup_cost}
        \centering
        \scalebox{.713}{
            \begin{tabular}{ccc|cc} 
                \toprule
                Cost & Loss & Angle Def. & AP$_{50}$ & w$\slash $ NMS \\ 
                \midrule
                L1 & L1 & Long Edge & 67.99 & +1.34 \\
                L1 & L1 & OpenCV & 69.26 &  +0.58 \\
                XYWH L1 & XYWH L1 & Long Edge & 69.88 & +0.04 \\
                XYWH L1 & L1 & Long Edge & 69.84 & +0.03 \\
                None & None & Long Edge & 68.30 & +0.10 \\
                None & L1 & Long Edge & 70.18 & -0.26 \\
                Hausdorff & L1 & Long Edge & \textbf{70.81} & -0.09 \\
                \bottomrule
            \end{tabular}}
    \end{minipage}
         \begin{minipage}[t]{0.24\linewidth}
         \subcaption{Denoising methods}
        \label{tab:noise}
        \centering
        \scalebox{0.68}{\begin{tabular}{c|c|c} 
                \toprule
                Method & Noise-Level & AP$_{50}$ \\ 
                \midrule
                Static & 0.2 & 76.20 \\
                Static & 0.4 & 75.63 \\
                Static & 0.6 & 76.24 \\
                Static & 0.8 & 75.53 \\
                Static & 1.0 & 76.14 \\
                \midrule 
                AQD & 1.0 & 78.06 \\
                AQD* & 1.0 & 78.68  \\
                \bottomrule
        \end{tabular}}
    \end{minipage} 
\end{table*}

\subsection{High-precision Evaluation}
\label{sec:high-precision}
Zeng et al. ~\cite{zeng2023ars} have underscored that AP$_{50}$ may not be the most appropriate metric for evaluating a model’s ability to accurately predict the orientation of objects, as it permits large angle deviations.
The authors argued that AP$_{75}$ is a more rigorous metric for assessing the performance in oriented object detection.
Table \ref{tab:dota1_ap75} presents a comparison of our model with other models in terms of AP$_{50}$ and AP$_{75}$ on the DOTA-v1.0.
\# points indicates the number of points for the Hausdorff distance.
Here, we also have conducted ablation studies to assess the impact of various IoU metrics $\mathcal{L}_\mathrm{iou}$, both in terms of matching cost and regression loss, on the performance of our model. Note that all these models are trained with the Hausdorff distance to replace the L1 distance matching cost.
Specifically, We compare our default IoU metric, the KLD loss~\cite{yang2021kld}, with the other alternatives like GWD~\cite{yang2021gwd}, Rotated IoU (RIoU)~\cite{zhou2019iou}, and GRIoU, the rotated variant of the generalized intersection over union~\cite{rezatofighi2019giou}.
The results indicate that our model excels not only in AP$_{50}$ but also in AP$_{75}$, outperforming other models, including those specifically pre-trained for accurate rotated representation.
Furthermore, our results indicate that using RIoU loss, including GRIoU, significantly enhances the performance in AP$_{75}$ compared to the models trained with metrics based on the 2D Gaussian distribution.

\subsection{Ablation Studies}
\label{sec:ablation}
Table \ref{tab:ablation_all} shows ablation experiments of the proposed module and other options. 
\textbf{Effect of each proposed component.} Table \ref{tab:components_ablation} demonstrates the performance improvement by proposed components.
Here, AQD and AQD* represent adaptive query denoising and \textit{improved adaptive query denoising}, outlined in Equation \ref{eq:dd} and Equation \ref{eq:improved_dd}, respectively.
The baseline is our implementation of the rotated Deformable DETR. Unless otherwise specified, we used the KLD loss for the IoU term in the matching cost and the training objective.
Applying the Hausdorff cost and the adaptive query denoising results in an increase in AP$_{50}$ to \textbf{78.05} and in AP$_{75}$ to \textbf{50.59}.
Moreover, as mentioned in Section \ref{sec:Adaptive Query Denoising via Bipartite Mathcing}, when \textit{improved adaptive query denoising} is applied, both AP$_{50}$ and AP$_{75}$ improve to \textbf{78.68} and \textbf{51.17}, respectively.
\textbf{Duplicate prediction comparison on patch-based validation according to $\mathcal{L}_\mathrm{L1}$.} 
We perform ablation studies on the DOTA-v1.0 patch validation set to assess the impact of the L1 loss term, $\mathcal{L}_\mathrm{L1}$, and explore alternative approaches to address the issue of duplicate predictions, as discussed in Section \ref{sec:bipartite}.
It is important to note that the issue of duplicate predictions is only observed in the patch validation set, as NMS postprocessing is typically applied in the final submission to the DOTA-v1.0 test set. Inference for the test set is generally performed on overlapping patches to handle large images within memory limitations.
Here, \textbf{None} means that we set $\lambda_\mathrm{L1} $, the weight for $\mathcal{L}_\mathrm{L1}$, to 0 and XYWH-L1 represents the L1 distance using $\{b_x, b_y, b_w, b_h\}$.
Table \ref{tab:dup_cost} summarizes the results of these ablation experiments.
The results indicate that replacing the L1 distance with the Hausdorff distance for the matching cost yields notable performance improvements and effectively reduces duplicate predictions.
While other metrics can be utilized to minimize duplicate predictions, the Hausdorff distance shows the best performance both with and without NMS.
\textbf{Effect of denoising methods.}
We stated that the high-level noisy queries in the later training phase hamper the training of the model. Instead of dynamically eliminating abundant queries, one can control the hyperparameters of the box noise level of the original denoising method. Table \ref{tab:noise} compares the results on DOTA-v1.0. \textbf{Static} refers to the original contrastive denoising training from DINO. It illustrates that dynamically filtering queries through bipartite matching is superior to merely adjusting noise hyperparameters. 
Ablation studies regarding the number of points for the Hausdorff distance and the method for updating the angle are presented in Appendix ~\ref{sec:more_able}.

\section{Conclusions}
This study presents a strong DETR that outperforms traditional oriented object detectors. Our comprehensive analysis demonstrates the square-like problem in bipartite matching from DETR series, which hinders non-duplicate predictions in DETR. To this end, we introduce the Hausdorff distance as a replacement for the L1 cost in bipartite matching.
Furthermore, we addresss the limitations of the static denoising approach in training rotated DETR models. Our adaptive query denoising technique, leveraging bipartite matching, effectively filters out irrelevant noised queries, thereby enhancing the training process and the model's overall performance.
The state-of-the-art performance of our RHINO on benchmark datasets demonstrates its significant potential for rotated object detection. This study contributes a strong DINO-based baseline for oriented object detection and underscores the potential of query-based object detectors, setting the stage for future research in this domain.

\newpage
\newpage
\appendix
\section*{Appendices}
\addcontentsline{toc}{section}{Appendices}
\renewcommand{\thesubsection}{\Alph{subsection}}

\subsection{Implemenation Details of Our Baseline}\label{sec:supple1_details}
In this section, we delve into the unique challenges associated with introducing rotation capabilities to DETR models and detail our specific implementation strategies.

\paragraph{Angle Prediction by Simple Extended Box Regressor.}
In the DETRs, the coordinates of objects are handled as normalized ranges [0, 1] (relative to the image size) in the training stage.
The normalized predicted coordinates from the $\mathrm{sigmoid}$ function are recovered to the original range by multiplying the width and height of the input image.
One straightforward method to equip DETRs with the capability to predict an object's orientation is by extending the output of the regression head to $\mathbf{b} \in [0, 1]^5$, where $\mathbf{b} = [b_x, b_y, b_w, b_h, b_{\theta}] $ represents the normalized center coordinates, box width and height, and the angle in radians, respectively.
Similar to the center coordinates and width and height, the angle should also be normalized.
To facilitate this, we adopt the Long Edge Definition $(\theta \in [0, \pi))$ and normalize angle values to lie within the [0, 1] range by dividing the ground truth angles by $\pi$. 
During the training phase, the rotated DETR is trained to predict these normalized coordinates.
At the inference time, the predictions are converted back to their original scale by multiplying the output $\mathbf{b}$ with $ [I_w, I_h, I_w, I_h, \pi ]$, where $I_w$ and $I_h$ denote the width and height of the image, respectively.
By treating angles the same way as other coordinate values, rather than introducing a dedicated angle prediction branch, the model can easily incorporate other developed components related to coordinates, such as iterative bounding box refinement, handling queries as anchor boxes, and proposing queries from the feature map of the encoder.

\paragraph{Replacement of GIoU Loss.}
The standard costs and losses used in DETRs are Focal loss, L1 loss, and GIoU loss.
However, calculating overlaps between two rotated objects is commonly known to be indifferentiable~\cite{yang2021gwd, yang2021kld, yang2022kfiou}.
Therefore, we need to consider alternative options to replace GIoU loss for rotated objects.
Two representative substitutes for representing the error between predictions and ground truth for rotated objects are Gaussian Wasserstein Distance (GWD)~\cite{yang2021gwd} and Kullback-Leibler Divergence (KLD)~\cite{yang2021kld}.
The center distance term in GWD and KLD also works as the penalty term for non-overlapping bounding boxes, similar to GIoU loss.
We adopt KLD as our baseline to replace the GIoU cost and loss since previous work has shown that KLD outperforms GWD in oriented object detection.
Additionally, we set the weight for the IoU term from 2 to 5 based on the comparison experiment, which is shown in Table \ref{tab:cost_weight}.
This weight is applied to all our experiments except the experiments conducted on the patch validation set.

\begin{figure}[b]
    \centering
    \begin{subfigure}{.31\textwidth}
      \centering
      \includegraphics[height=5cm]{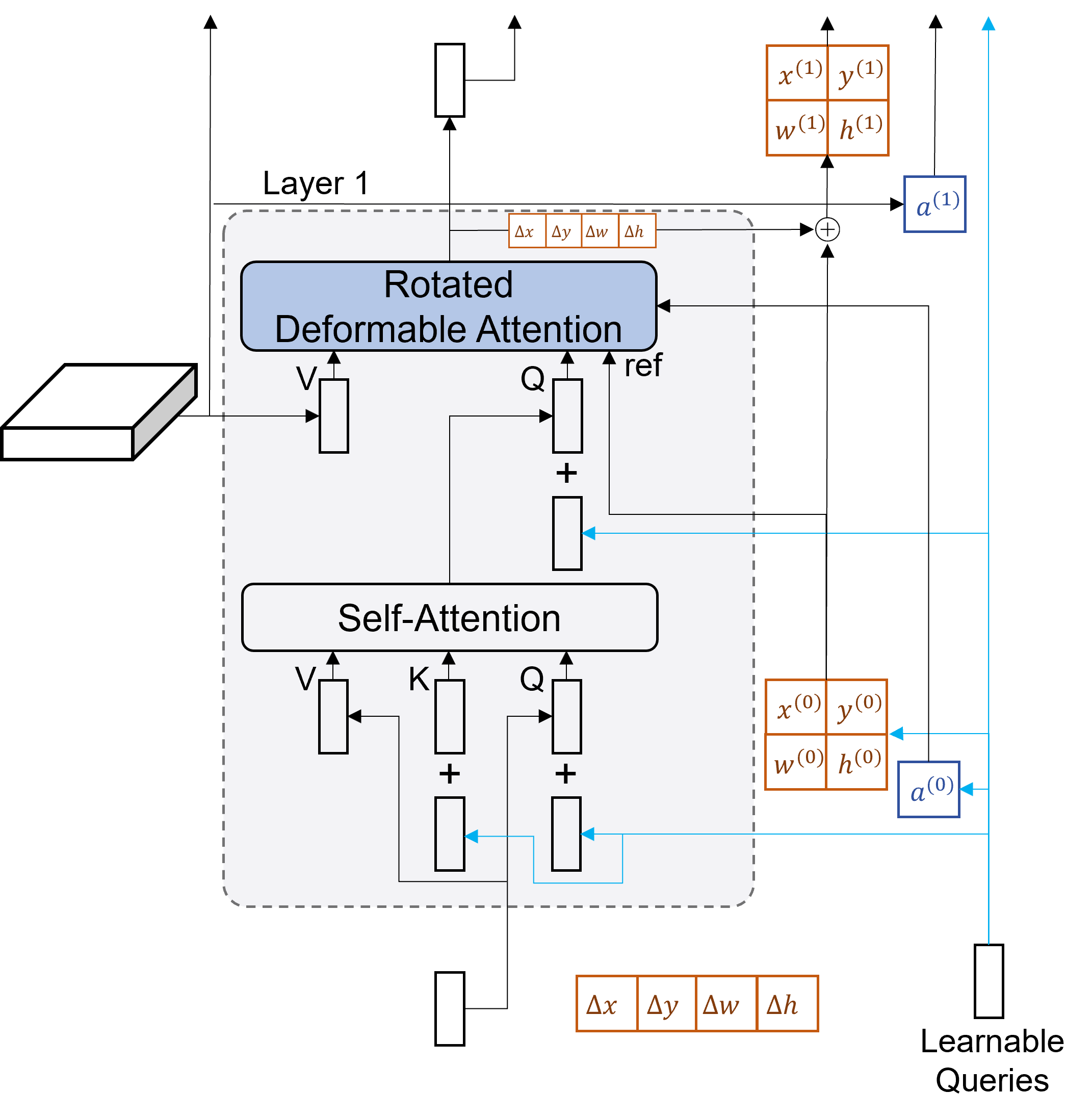}
      \caption{ARS-DETR}
      \label{fig:ars_detr}
    \end{subfigure}
    \hspace{10pt}
    \begin{subfigure}{.31\textwidth}
      \centering
      \includegraphics[height=5cm]{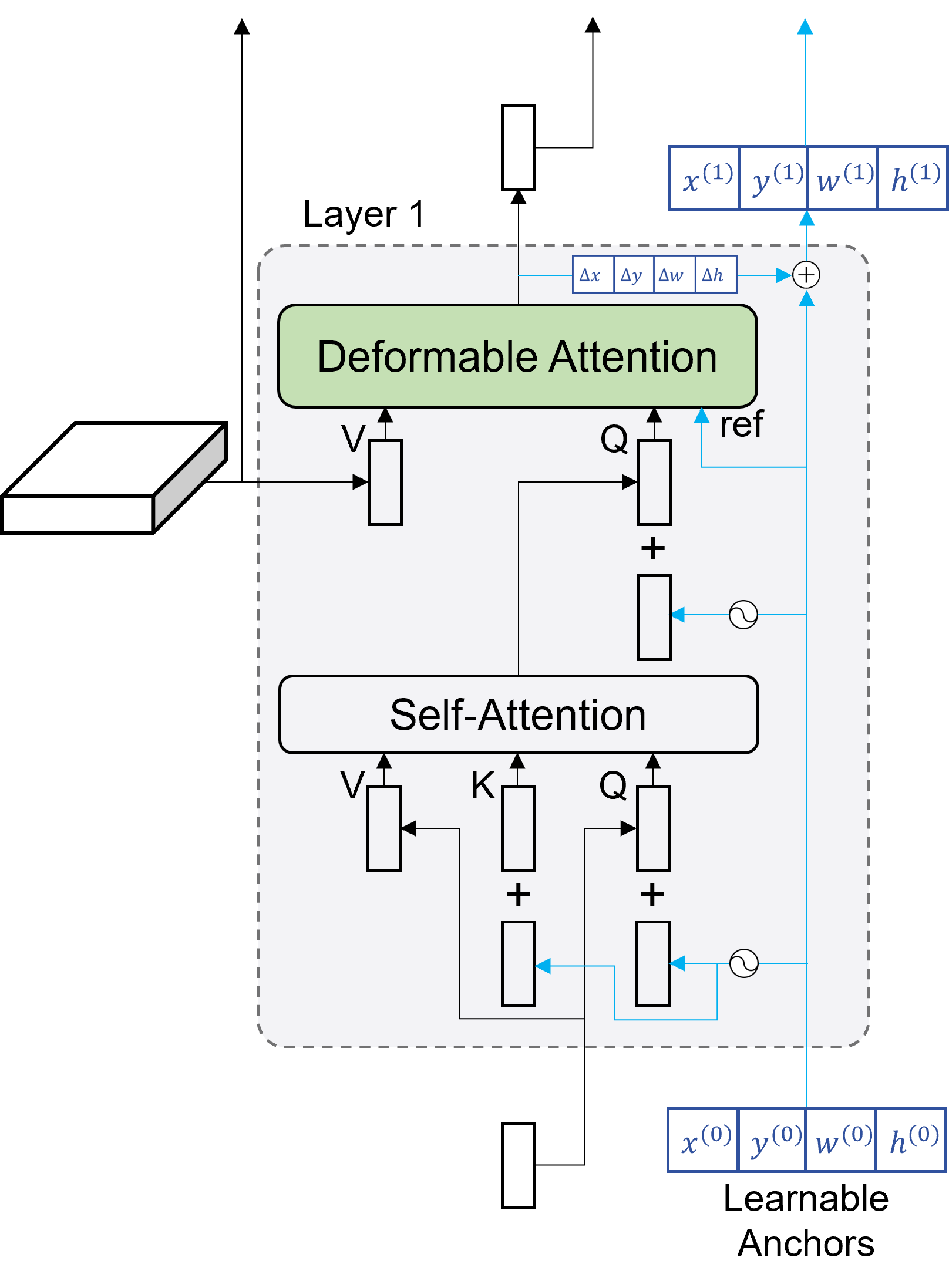}
      \caption{DAB-DETR}
      \label{fig:dab_detr}
    \end{subfigure}
    \begin{subfigure}{.31\textwidth}
      \centering
      \includegraphics[height=5cm]{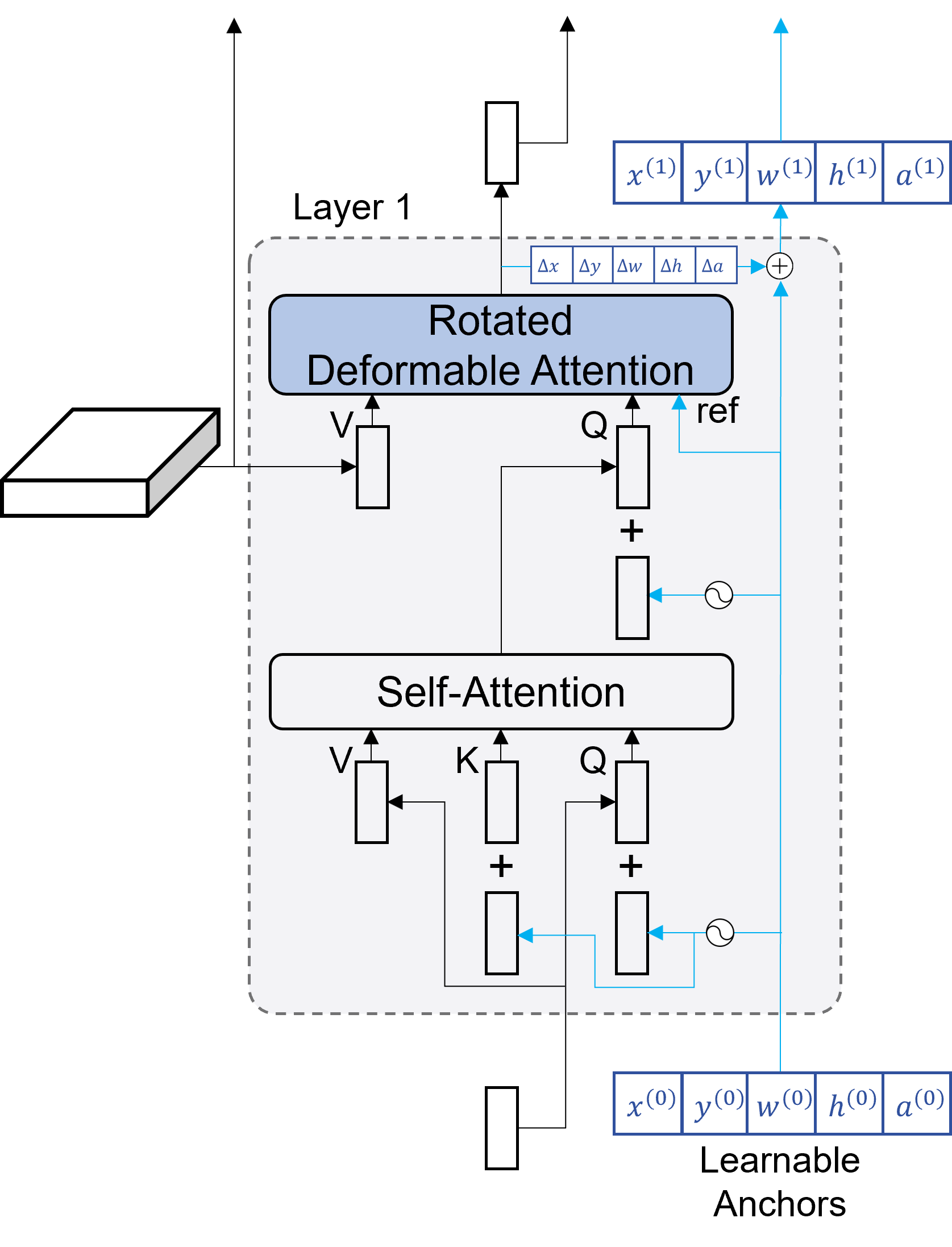}
      \caption{Ours}
      \label{fig:rot_deform_attn}
    \end{subfigure}
    \caption{Comparison of attention layers in different models. (a) ARS-DETR, based on Deformable DETR, predicts reference boxes from learnable queries but does not fully integrate angle information into the objects' spatial information. (b) DAB-DETR, which our model builds upon, defines learnable queries as horizontal anchors. (c) Our model extends this approach by defining anchors as rotated boxes, leveraging methods such as iterative refinement and two-stage proposals from DINO, resulting in improved integration of angle information.}
    \label{fig:attentions}
\end{figure}

\paragraph{Rotated Deformable Attention.}
In order to address the slow convergence and high computational memory complexity associated with the multi-head attention module in the original DETR, Zhu et al. \cite{zhu2020deformable} proposed a deformable attention module.
This module takes an input feature map and either 2D reference points $\mathbf{p_2} = (p_x, p_y)$ or 4D reference boxes $\mathbf{p_4} = (p_x, p_y, p_w, p_h)$, where $(p_x, p_y)$ represents a series of center coordinates and $(p_w, p_h)$ represents a series of widths and heights for the reference boxes, to determine the sampling location.
The sampling locations for 2D reference points are obtained by adding the predicted sampling offsets to the given reference points.
In the case of 4D reference boxes, the predicted sampling offsets are first added to the center coordinates of reference boxes.
Then, the resulting values are multiplied by half of the widths and heights of the reference boxes to determine the final 2D coordinates for the sampling locations.
Using these 4D reference boxes enhances performance through iterative refinement and query proposal from the encoder.

As Zhu et al.~\cite{zeng2023ars} pointed out, using $\{b_x, b_y, b_w, b_h\}$ as reference boxes from 5D reference boxes by dropping $b_\mathrm{rad}$ leads to feature misalignment, especially in deformable attention modules.
We apply a similar way to ARS-DETR~\cite{zeng2023ars} to rotate the center coordinates from the reference boxes.
The predicted normalized angles are recovered by multiplying the normalized factor $\pi$, and the next steps follow the case of 4D reference boxes as described previously.
In contrast to the approach proposed by Zeng et al.~\cite{zeng2023ars}, we update 5D reference boxes including angles during the iterative refinement step.
To update the 5D reference boxes, we apply the $\mathrm{inverse~sigmoid}$ function to the reference points and add the resulting values to the logit of the regression head in logit space.
We then normalize the updated values using the $\mathrm{sigmoid}$ function, which maps them to the range of [0, 1].
The same update logic is also applied to the predicted regression in the head part.
A visualization of our deformable attention is illustrated in Figure~\ref{fig:attentions}. Building on the principles of DAB-DETR~\cite{liu2022dab}, which defines learnable queries as dynamic anchor boxes, our model integrates angles into oriented reference boxes in a natural way.

\paragraph{Extension to DINO.}
\label{sec:ext_dino}
By straightforwardly extending the regression head with 5D rotated boxes, the model is able to naturally adopt the developed mechanisms such as queries as dynamic anchor boxes proposed by successive works~\cite{liu2022dab, li2022dn, zhang2022dino}.
Here, we just brief some implementation details to note.
The two-stage scheme introduced by Deformable-DETR~\cite{zhu2020deformable}, requires \textbf{generating grid proposals from the encoded features}, and predicting query proposals by running the regression head to the encoded features. 
In our implementation, We use zero-angle grids to generate grid proposals from the encoded features and run the regression head on the flattened feature vectors. This allows the regression output to be added directly to the proposals.
For \textbf{denoising training}~\cite{li2022dn, zhang2022dino}, we add noises only to $\{b_x, b_y, b_w, b_h\}$ for simplicity.
We exclude the angle for the decoder positional queries which are related to \textbf{mixed query selection} in DINO~\cite{zhang2022dino}.
Finally, we experiment with models using 900 queries and set the maximum number of predictions to 500 to handle densely packed objects in remote sensing images.

\subsection{Implementation Details of RHINO}\label{sec:imple_rhino}
We implement our models on MMRotate~\cite{zhou2022mmrotate} v1.0.0rc1.
Most of our experiments were conducted on 2 NVIDIA V100 or A100 GPUs, with a total batch size of 8.
Two backbones: ResNet-50~\cite{he2016resnet} and Swin-Tiny~\cite{liu2021swin} are used in our experiments, both of which are pre-trained on ImageNet-1k~\cite{deng2009imagenet}.
We use the AdamW~\cite{kingma2014adam, loshchilov2017adamw} optimizer with an initial learning rate of $1 \times 10^{-4}$.
We use a step decay learning rate schedule where the learning rate is multiplied by 0.1 at the 11th epoch during the 12 epoch training and multiplied by 0.1 again at the 27th and 33rd epochs during the 36 epoch training.
For DOTA-v1.0/v1.5/v2.0, we crop each image in the data sets to $1024 \times 1024$ pixels with a 200-pixel overlap.
The input size of experiments on DIOR-R is set to $800 \times 800$.
We use only horizontal, vertical, and diagonal flips without additional augmentations.

\subsection{More Ablation Studies on Model Components.}\label{sec:more_able}
\begin{figure}
    \centering
    \begin{subfigure}{.48\columnwidth}
      \centering
      \includegraphics[height=4cm]{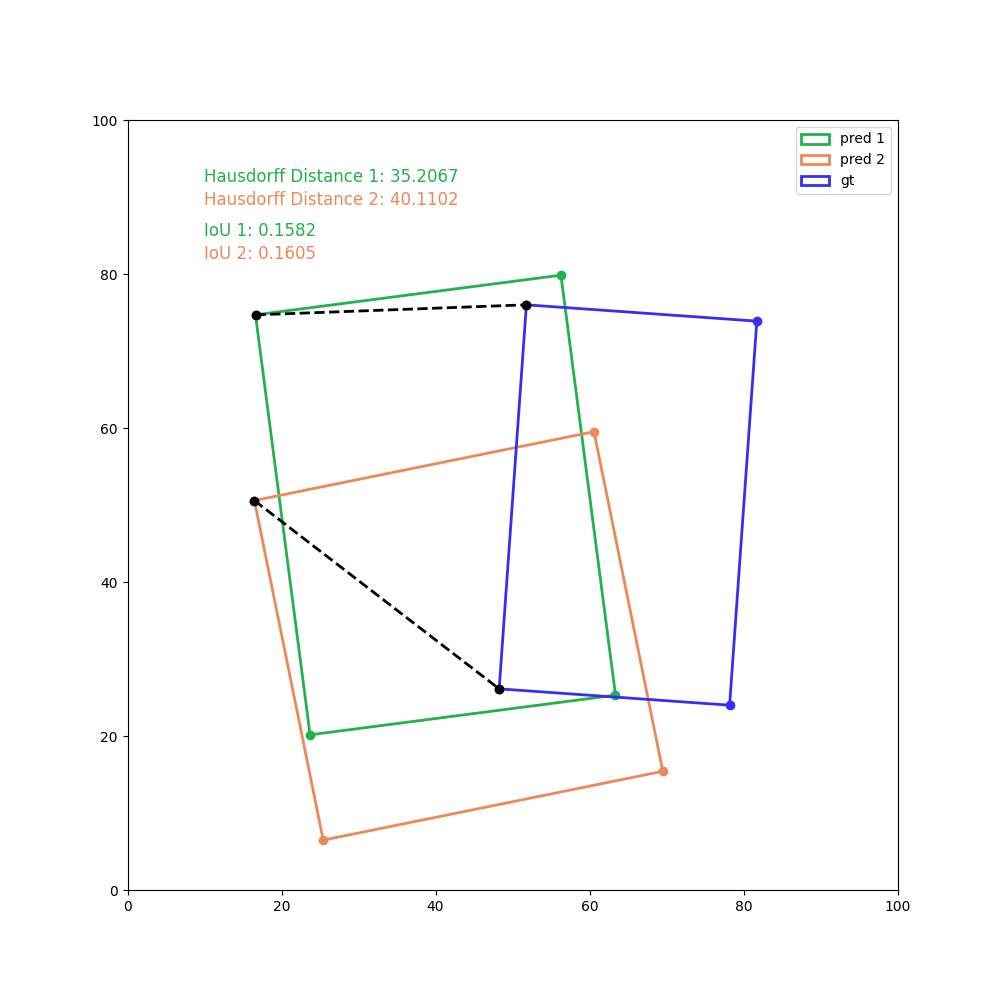}
      \subcaption{Hausdorff distance with 4 points.}
      \label{fig:hausdorff_4_points}
    \end{subfigure}
    \hfill
    \begin{subfigure}{.48\columnwidth}
      \centering
      \includegraphics[height=4cm]{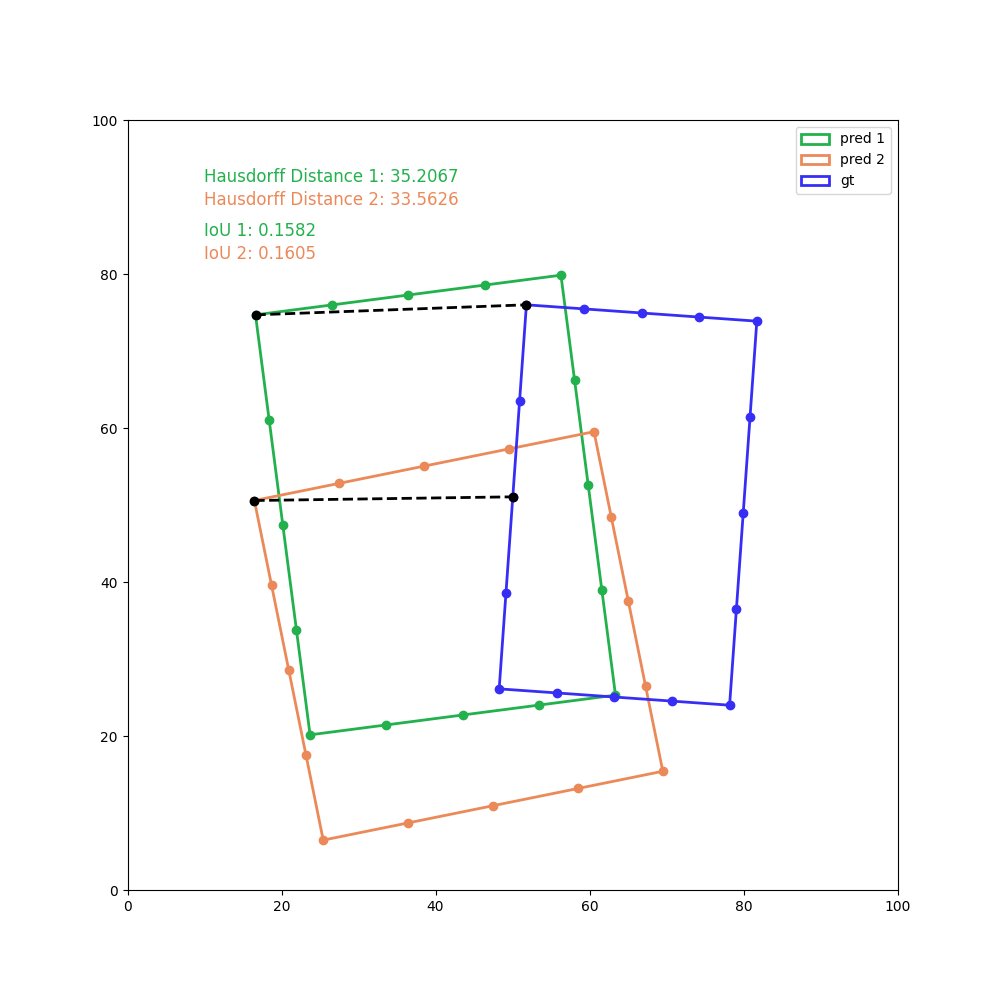}
      \subcaption{Hausdorff distance with 16 points.}
      \label{fig:hausdorff_16_points}
    \end{subfigure}
    \caption{Visualization of the Hausdorff distance for different numbers of points. Matching is based on the smaller Hausdorff distance, with higher IoU indicating better matching.}
    \label{fig:hausdorff_comparison}
\end{figure}

\textbf{Effect of the number of points for the Hausdorff distance.}
In our approach, we approximate the Hausdorff distance between two rotated boxes using points along the box edges. As illustrated in Figure \ref{fig:hausdorff_comparison}, increasing the number of points improves the precision of matching, as reflected in the IoU.
To assess the effect of point count, we evaluated different configurations, with the results summarized in Table \ref{tab:num_points}. The 4-point configuration achieved the highest performance in AP${50}$, with a score of 78.68. However, the 32-point configuration yielded the best result in AP${75}$, achieving 51.84. Since AP${75}$ uses a stricter IoU threshold of 0.75 compared to 0.5 in AP${50}$, this indicates that increasing the number of points improves performance in more precise matching evaluations. Nevertheless, for simplicity and efficiency, we use 4 points to compute the Hausdorff distance in all other experiments.

\begin{table*}
\caption{Further ablation studies on DOTA-v1.0.}
\label{tab:ablation_all2}
    \centering
    \begin{minipage}{.45\linewidth}
            \subcaption{Robustness on weight for IoU term}
            \label{tab:cost_weight}
            \centering
            \scalebox{0.8}{\begin{tabular}{ccc|c|c} 
                \toprule
                $\mathcal{L}_\mathrm{L1}$ Cost & $\mathcal{L}_\mathrm{L1}$ Loss &  $\lambda_\mathrm{iou}$ & AP$_{50}$ & w\/ NMS  \\ 
                \midrule
                None & None & 2 & 68.30 & +0.10  \\
                None & None & 5 & 70.18 & -0.05  \\
                Hausdorff & L1 & 2 & 70.81 & -0.09  \\
                Hausdorff & L1 & 5 & 70.94 & +0.16  \\
                \bottomrule
            \end{tabular}}
    \end{minipage}
    \hspace{0.5cm}
    \begin{minipage}{.45\linewidth}
            \subcaption{The number of points for the Hausdorff distance}
        \label{tab:num_points}
        \centering
        \scalebox{0.8}{
            \begin{tabular}{cc|cc} 
                \toprule
               \# points & AQD*  & AP$_{50}$  & AP$_{75}$ \\ 
                \midrule
                4 & \checkmark & 78.68 & 51.17 \\ 
                8 & \checkmark & 78.12 & 51.02  \\
                32 & \checkmark & 78.49 & 51.84 \\
    
                \bottomrule
            \end{tabular}}
    \end{minipage}
     \begin{minipage}{0.8\linewidth}
         \subcaption{Effect of updating angles in the baseline}
        \label{tab:baseline_abl}
    \centering
    \scalebox{0.8}{
            \begin{tabular}[b]{l|c|cccc|c} 
                \toprule
                Model & Epochs & IR & Two-Stage &  5D RP & 5D Head & AP$_{50}$ \\ 
                Deformable DETR & 50 & - & - & - & - & 68.50 \\
                Deformable DETR & 50 & \checkmark & -  & - &
                -& 68.54 \\
                Deformable DETR & 50 & \checkmark & \checkmark & - & - & 70.48 \\
                \midrule
                DINO & 12 & \checkmark & \checkmark & - & - & 71.36 \\
                DINO & 12 & \checkmark & \checkmark & \checkmark & - & 72.76 \\
                DINO & 12 & \checkmark & \checkmark & \checkmark & \checkmark & 76.10 \\
                \bottomrule
                \bottomrule
    \end{tabular} }
    \end{minipage} 
\end{table*}

\textbf{Updating angles in logit space.}
Table \ref{tab:baseline_abl} presents the importance of updating angles in DETR components.
\textbf{IR} and \textbf{Two-Stage} represent iterative bounding box refinement and a two-stage approach, respectively, as introduced by Deformable DETR, but without incorporating the angle parameter. \textbf{5D RP} refers to the updating angles of 5D reference points in Decoder layers, while \textbf{5D Head} refers to the updating angles of predictions using reference points in the head module. No checkmark in \textbf{5D RP} or \textbf{5D Head} columns indicates that only the coordinates $\{ b_x, b_y, b_w, b_h\}$ of the reference points or predictions are updated.
This finding illustrates that treating oriented object detection as a regression problem benefits from adopting the latest query-as-bounding-box approaches proposed in state-of-the-art DETR models.

\textbf{Analysis of the impact of aspect ratios.}
To evaluate the effectiveness of the Hausdorff distance in addressing the issue of duplicate predictions, we conducted a class-wise performance comparison using the DOTA-v1.0 patch validation set.
The results, detailed in Table \ref{tab:aspect}, reveal that using the Hausdorff distance significantly improves performance for square-like objects, such as {\tt baseball-diamond}, {\tt storage-tank}, and {\tt roundabout}.

\begin{table*}[t]
        \centering
        \caption{Computational cost comparison using 2 A-40 GPUs on DOTA-v1.0}
            \label{tab:resource}
        \scalebox{0.9}{
            \begin{tabular}{l|r|r|c|c|c|c|c} 
                \toprule
                Method &  Params & GFLOPs & FPS$_{bs=1}$ & Inf. Memory & Train Time & Train  Memory & AP$_{50}$ \\ 
                \midrule
                DCFL~\cite{xu2023coarsetodynamic} & 36.1M & 157.80 & 23 & 2.3 GB & 11 Hours & 4 GB & 75.35 \\
                ARS-DETR~\cite{zeng2023ars} & 41.6M & 205.95 & 14 & 2.2 GB & 24 Hours & 10-11 GB & 73.42 \\
                \midrule
                DINO & 47.3M & 280.38& 13 & 2.2 GB & 25 Hours & 16-20 GB & 74.56 \\
                + Hausdorff dist. & 47.3M & 280.38 & 13 & 2.2 GB & 25 Hours & 16-20 GB & 76.14  \\
                + AQD* & 47.3M & 280.38 & 13 & 2.2 GB & 30 Hours &  16-20 GB & 78.68  \\
                \bottomrule
            \end{tabular}
            }
    \label{tab:computation}
\end{table*}

\begin{table}
    \caption{Class-wise performance and its average aspect ratios on the patch validation set.}
    \label{tab:aspect}
        \centering
        \scalebox{0.7}{
            \begin{tabular}{l|c|c|c} 
                \toprule
                class & Aspect Ratio & L1 & Hausdorff \\ 
                \midrule
                plane & 0.7982 & 89.0 & 89.8 \\
                baseball-diamond & \textbf{0.9239} & \textbf{67.1} & \textbf{71.5} \\
                bridge & 0.4350 & 48.6 & 50.9 \\
                ground-track-field & 0.5044 & 66.3 & 66.8 \\
                small-vehicle & 0.4736 & 69.2 & 69.1 \\
                large-vehicle & 0.2746 & 84.5 & 83.7 \\
                ship & 0.3535 & 87.7 & 88.4 \\
                tennis-court & 0.4828 & 90.6 & 90.5 \\
                basketball-court & 0.5748 & 60.0 & 56.7 \\
                storage-tank & \textbf{0.9365} & \textbf{63.3} & \textbf{73.1} \\
                soccer-ball-field & 0.6109 & 56.2 & 58.2 \\
                roundabout & \textbf{0.9224} & \textbf{52.3} & \textbf{60.8} \\
                harbor & 0.4018 & 75.8 & 76.5 \\
                swimming-pool & 0.6087 & 53.7 & 54.9 \\
                helicopter & 0.7146 & 55.5 & 71.4 \\
                \midrule
                AP$_{50}$  & 0.4705 & 68.0 & 70.81 \\
                \bottomrule
            \end{tabular}}
\end{table}

\begin{table}
        \centering
            \caption{Performance comparison on rotated object datasets}
            \label{tab:other_datasets}
        \scalebox{0.8}{
            \begin{tabular}{c|c|cc} 
                \toprule
                Method & Dataset & AP$_{50}$ & Hmean$_{50}$ \\ 
                \midrule
                DINO & \multirow{2}{*}{SKU110K-R}  & 87.22 & - \\
                \ours RHINO & & \ours 88.30 & -  \\
                \midrule
                \midrule
                DINO & \multirow{2}{*}{MSRA-TD500} & - & 0.5862  \\
                \ours RHINO & & - & \ours 0.6173  \\
                \midrule
                DINO & \multirow{2}{*}{ICDAR2015} & - & 0.5685  \\
                \ours RHINO & & - & \ours 0.5819  \\
                \bottomrule
            \end{tabular}
            }

\end{table}

\begin{table}
            \caption{Performance comparison on HRSC2016.}
            \label{tab:hrsc}
        \centering
        \scalebox{0.8}{
            \begin{tabular}{l|c|c|c} 
                \toprule
                Method & Backbone & mAP$_{07}$ & mAP$_{12}$ \\ 
                \midrule
                RoI Trans. \cite{ding2019roitrans} & R-101 & 86.20 & - \\
                R$^3$Det \cite{yang2021r3det} & R-101 & 89.26 & 96.01 \\
                S$^2$ANet \cite{han2021align} & R-101 & 90.17 & 95.01 \\
                Oriented R-CNN \cite{xie2021orientedrcnn} & R-50 & 90.40 & 96.50 \\
                ReDet \cite{han2021redet} & ReR-50 & 90.46 & 97.63 \\
                RTMDet-R-tiny \cite{lyu2022rtmdet} & CSPNeXt-tiny & \textbf{90.60} & 97.10 \\
                \midrule
                 DINO \cite{zhang2022dino} (Our implementation) &  R-50 &  90.26 &  97.37 \\
                \ours RHINO (Ours) & \ours R-50 & \ours 90.30 &  \ours \textbf{97.86} \\
                \bottomrule
            \end{tabular}
            }
\end{table}

\subsection{Computational Cost Comparison}
\label{sec:cost-speed}
Table \ref{tab:resource} presents a performance comparison of different methods. Note that DINO requires more computation due to its Transformer modules. Using the Hausdorff distance (with 4 corner points) enhances performance without significantly increasing computation. The adaptive query denoising method incurs higher computational costs because it involves additional bipartite matching during the denoising process. However, the inference speed and number of parameters remain unchanged compared to DINO, as the proposed methods primarily affect model training.

\subsection{Experiments on Other Rotated Object Datasets}
We extend the evaluation of our model, RHINO, by comparing it with the DINO baseline across various other rotated object detection datasets. Specifically, we focus on two scene text detection datasets, MSRA-TD500 and ICDAR2015, as well as the retail object dataset SKU110K-R.
For the MSRA-TD500 dataset, we trained both RHINO and DINO models for 50 epochs. For the SKU110K-R and ICDAR2015 datasets, the training duration was set to 36 epochs. As detailed in Table \ref{tab:other_datasets}, RHINO consistently outperforms the DINO baseline, demonstrating its robustness and effectiveness even though these datasets generally include non-square objects.

Table \ref{tab:hrsc} presents a comparison between our model and state-of-the-art models on the HRSC2016, which is a widely used aerial ship dataset for oriented object detection.
While our model, RHINO, exhibits marginally lower performance than the state-of-the-art models in terms of mAP$_{07}$, it is noteworthy that RHINO surpasses all other models in mAP$_{12}$.

\subsection{Visualization}
\begin{figure*}
    \centering
    \includegraphics[width=0.9\linewidth]{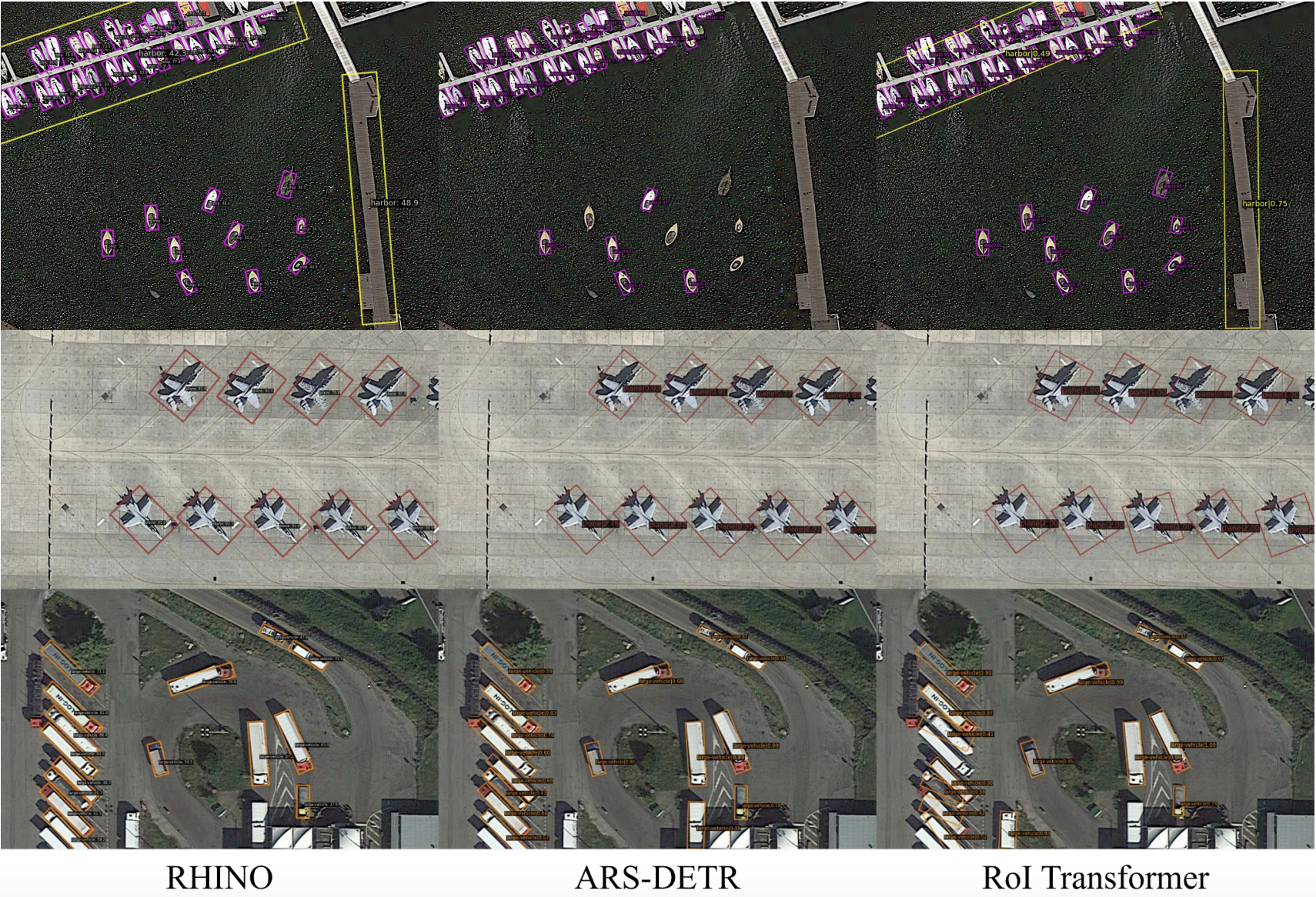}
    \caption{Qualitative comparison between our model and other models on the DOTA-v1.0 dataset.}
    \label{fig:vis_dota}
\end{figure*}

Figure \ref{fig:vis_dota} presents a qualitative comparison of our models and other models.
Notably, the RoI Transformer demonstrates lower precision in predicting object angles compared to our model and ARS-DETR.

\begin{figure*}
    \centering
    \includegraphics[width=0.9\linewidth]{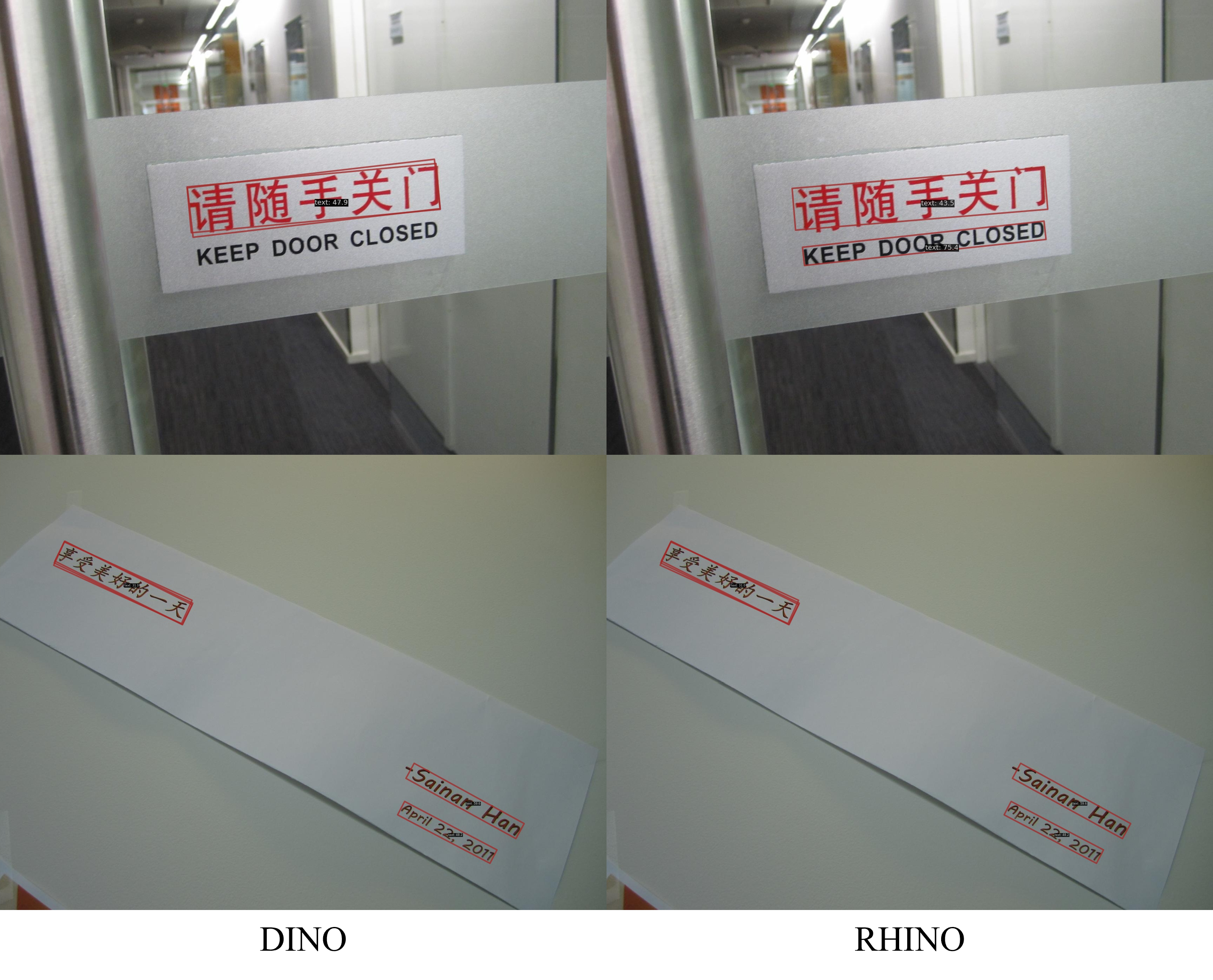}
    \caption{Qualitative comparison between the baseline and our model on the MSRA-TD500 dataset.}
    \label{fig:vis_msra}
\end{figure*}

\begin{figure*}
    \centering
    \includegraphics[width=0.9\linewidth]{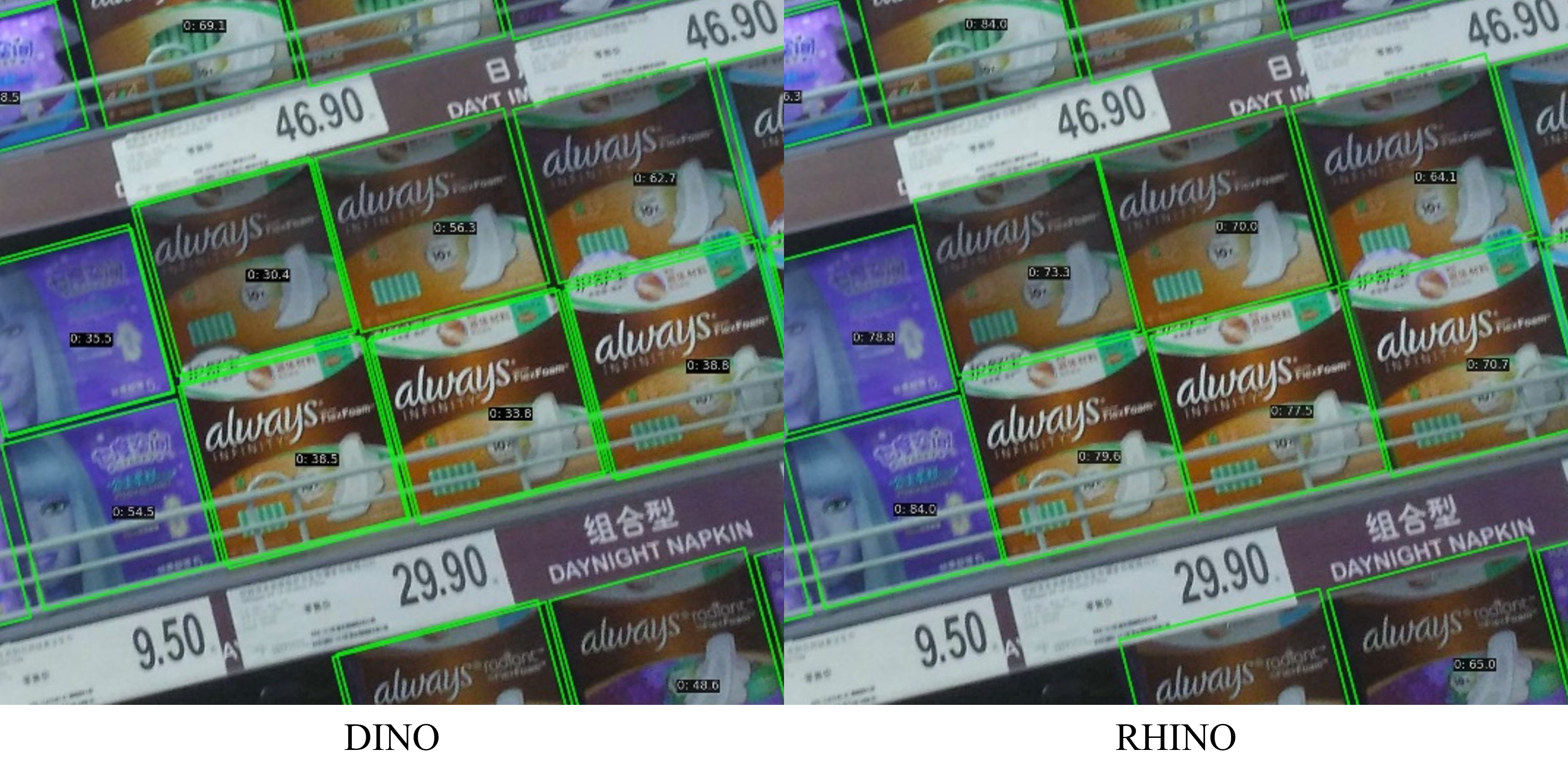}
    \caption{Qualitative comparison between the baseline and our model on the SKU110K-R dataset.}
    \label{fig:vis_sku}
\end{figure*}

Figure \ref{fig:vis_msra} and Figure \ref{fig:vis_sku}  provide further comparisons between the baseline and our model.
Overall, RHINO exhibits relatively consistent non-duplicate predictions compared to the baseline, especially in square-like objects, as shown in Figure \ref{fig:vis_sku}.
However, it is important to note that RHINO occasionally produces duplicate predictions on the MSRA-TD500 dataset.
We assume this is due to the small amount of training dataset for MSRA-TD500.

\subsection{Limitation}\label{sec:limitations}
Despite the significant performance improvements achieved by our model, some limitations exist. The adaptive query denoising method, which leverages bipartite matching to filter out harmful noised queries selectively, necessitates extended training times, as shown in Table~\ref{tab:resource}. Furthermore, while adaptive query denoising consistently enhances performance across various rotated object detection tasks, its application to the DINO model trained on the COCO dataset results in a slight performance decrement (-1.6 AP) compared to the baseline. We suspect this reduction in performance may stem from the observation that the denoising task for horizontally aligned objects is less impacted by noisy queries, unlike their rotated counterparts.
Future work may explore more improved methods to adaptively control denoising tasks, potentially bypassing the need for bipartite matching. such as the adaptive matching cost for denoising. This might include developing an adaptive matching cost specifically tailored for the denoising task, which could offer a more nuanced approach to improving model performance while mitigating extended training times.

\begin{algorithm*}
\caption{Contrastive Query Denoising of DINO ~\cite{zhang2022dino}.}
\begin{algorithmic}[1]
    \STATEx \textbf{Input:} Predicted objects $\mathbf{p}=\{p_0, p_1, \dots, p_{N-1} \}$, Ground truths $\mathbf{y}=\{y_0, y_1, \dots, y_{M-1}\}$, Positive noised queries $\mathbf{q}^\mathrm{pos}=\{q^\mathrm{pos}_0, q^\mathrm{pos}_1, \dots, q^\mathrm{pos}_{M-1} \}$, Negative noised queries $\mathbf{q}^\mathrm{neg}=\{q^\mathrm{neg}_0, q^\mathrm{neg}_1, \dots, q^\mathrm{neg}_{M-1} \}$.
    \STATEx 
    \STATE \textbf{Initialization:} $\mathbf{p}^\mathrm{pos}=f_\mathrm{refine}(\mathbf{q}^\mathrm{pos})$, $\mathbf{p}^\mathrm{neg}=f_\mathrm{refine}(\mathbf{q}^\mathrm{neg})$.
    
    \STATEx \textbf{Calculate the loss:} $\mathcal{L}_\mathrm{denoising}(\mathbf{p}^\mathrm{pos}, \mathbf{p}^\mathrm{neg}, \mathbf{y}) = \sum_{i=0}^{M-1} \mathcal{L}_\mathrm{pos}(p^\mathrm{pos}_{i}, y_i)+ \mathcal{L}_\mathrm{neg}(p_i^\mathrm{neg}, \varnothing)$.
    
    \STATEx \textbf{Match Queries and Predictions}
    \State Set assignment $\sigma$ with the corresponding ground truths.
    \State $\sigma = {0, 1, \dots, M-1}$.
    
    \Statex \textbf{Filter Unhelpful Queries:}
    \State $\mathcal{L}_\mathrm{pos}(p^\mathrm{pos}_{i}, y_i) = \mathbbm{1}_{\{\sigma_i = i\}}\mathcal{L}_\mathrm{train}(p^\mathrm{pos}_{i}, y_i)$.    
    \State \textbf{Update:} Perform backpropagation and update model parameters.
    
\end{algorithmic}
\label{algo:cqd}
\end{algorithm*}

\begin{algorithm*}
\caption{Our Improved Adaptive Query Denoising. The \textcolor{blue} {blue text lines} represent the modifications and highlight the differences from Algorithm ~\ref{algo:cqd}.}
\begin{algorithmic}[1]
    \STATEx \textbf{Input:} Predicted objects $\mathbf{p}=\{p_0, p_1, \dots, p_{N-1} \}$, Ground truths $\mathbf{y}=\{y_0, y_1, \dots, y_{M-1}\}$, Positive noised queries $\mathbf{q}^\mathrm{pos}=\{q^\mathrm{pos}_0, q^\mathrm{pos}_1, \dots, q^\mathrm{pos}_{M-1} \}$, Negative noised queries $\mathbf{q}^\mathrm{neg}=\{q^\mathrm{neg}_0, q^\mathrm{neg}_1, \dots, q^\mathrm{neg}_{M-1} \}$.
    \STATEx 
    \STATE \textbf{Initialization:} $\mathbf{p}^\mathrm{pos}=f_\mathrm{refine}(\mathbf{q}^\mathrm{pos})$, $\mathbf{p}^\mathrm{neg}=f_\mathrm{refine}(\mathbf{q}^\mathrm{neg})$.
    
    \STATEx \textbf{Denoising Loss:} $\mathcal{L}_\mathrm{denoising}(\mathbf{p}^\mathrm{pos}, \mathbf{p}^\mathrm{neg}, \mathbf{y}) = \sum_{i=0}^{M-1} \mathcal{L}_\mathrm{pos}(p^\mathrm{pos}_{i}, y_i)+ \mathcal{L}_\mathrm{neg}(p_i^\mathrm{neg}, \varnothing)$.
    
    \STATEx \textbf{Match Queries and Predictions}
    \State \textcolor{blue} {$\mathbf{p}^\mathrm{all} = [\mathbf{p}^\mathrm{pos};\mathbf{p}]
    = [p^\mathrm{pos}_0, p^\mathrm{pos}_1, \dots, p^\mathrm{pos}_{M-1}, p_0, p_1, \dots, p_{N-1}]$.}
    \State Find optimal assignment $\sigma$ using bipartite matching.
    \State \textcolor{blue} { $\hat{\sigma} =\argmin_{\sigma \in \mathfrak{S}}\sum_{i=0}^{M+N-1} \mathcal{L}_\mathrm{match}(\hat{y}_i, \mathbf{p}^\mathrm{all}_{\sigma_i})$. }
    
    \Statex \textbf{Filter Unhelpful Queries:}
    \State \textcolor{blue} {$\mathcal{L}_\mathrm{pos}(p^\mathrm{pos}_{i}, y_i) = \mathbbm{1}_{\{\hat{\sigma}_i = i\}}\mathcal{L}_\mathrm{train}(p^\mathrm{pos}_{i}, y_i) + \mathbbm{1}_{\{\hat{\sigma}_i \neq i\}}\mathcal{L}_\mathrm{neg}(p^\mathrm{pos}_{i}, \varnothing)$.}
    
    \State \textcolor{blue} {$\mathcal{L}^*_\mathrm{pos}(p^\mathrm{pos}_{i}, y_i) = \mathcal{L}_\mathrm{pos}(p^\mathrm{pos}_{i}, y_i) + \mathbbm{1}_{\{\hat{\sigma}_i \neq i\}} \mathcal{L}_\mathrm{bbox}(p^\mathrm{pos}_{i},y_i)$.}
    
    \State \textbf{Update:} Perform backpropagation and update model parameters.
    
\end{algorithmic}
\label{algo:aqd}
\end{algorithm*}



{\small
\begin{thebibliography}{58}
\providecommand{\natexlab}[1]{#1}
\providecommand{\url}[1]{\texttt{#1}}
\expandafter\ifx\csname urlstyle\endcsname\relax
  \providecommand{\doi}[1]{doi: #1}\else
  \providecommand{\doi}{doi: \begingroup \urlstyle{rm}\Url}\fi

\bibitem[Xia et~al.(2018)Xia, Bai, Ding, Zhu, Belongie, Luo, Datcu, Pelillo, and Zhang]{xia2018dota}
Gui-Song Xia, Xiang Bai, Jian Ding, Zhen Zhu, Serge Belongie, Jiebo Luo, Mihai Datcu, Marcello Pelillo, and Liangpei Zhang.
\newblock Dota: A large-scale dataset for object detection in aerial images.
\newblock In \emph{Proceedings of the IEEE conference on computer vision and pattern recognition}, pages 3974--3983, 2018.

\bibitem[Ding et~al.(2021)Ding, Xue, Xia, Bai, Yang, Yang, Belongie, Luo, Datcu, Pelillo, and Zhang]{ding2021object}
Jian Ding, Nan Xue, Guisong Xia, Xiang Bai, Wen Yang, Micheal~Ying Yang, Serge~J. Belongie, Jiebo Luo, Mihai Datcu, Marcello Pelillo, and L.~Zhang.
\newblock Object detection in aerial images: A large-scale benchmark and challenges.
\newblock \emph{IEEE transactions on pattern analysis and machine intelligence}, 44\penalty0 (11):\penalty0 7778--7796, 2021.

\bibitem[Ding et~al.(2019)Ding, Xue, Long, Xia, and Lu]{ding2019roitrans}
Jian Ding, Nan Xue, Yang Long, Gui-Song Xia, and Qikai Lu.
\newblock Learning roi transformer for oriented object detection in aerial images.
\newblock In \emph{2019 IEEE/CVF Conference on Computer Vision and Pattern Recognition (CVPR)}, pages 2844--2853, 2019.
\newblock \doi{10.1109/CVPR.2019.00296}.

\bibitem[Yang et~al.(2023{\natexlab{a}})Yang, Yan, Liao, Yang, Tang, and He]{9756223}
X.~Yang, J.~Yan, W.~Liao, X.~Yang, J.~Tang, and T.~He.
\newblock Scrdet++: Detecting small, cluttered and rotated objects via instance-level feature denoising and rotation loss smoothing.
\newblock \emph{IEEE Transactions on Pattern Analysis and Machine Intelligence}, 45\penalty0 (02):\penalty0 2384--2399, feb 2023{\natexlab{a}}.
\newblock ISSN 1939-3539.
\newblock \doi{10.1109/TPAMI.2022.3166956}.

\bibitem[Ming et~al.(2021)Ming, Zhou, Miao, Zhang, and Li]{Ming_Zhou_Miao_Zhang_Li_2021}
Qi~Ming, Zhiqiang Zhou, Lingjuan Miao, Hongwei Zhang, and Linhao Li.
\newblock Dynamic anchor learning for arbitrary-oriented object detection.
\newblock \emph{Proceedings of the AAAI Conference on Artificial Intelligence}, 35\penalty0 (3):\penalty0 2355--2363, May 2021.
\newblock \doi{10.1609/aaai.v35i3.16336}.

\bibitem[Koo et~al.(2018)Koo, Seo, Jeon, Choe, and Jeon]{10.1145/3274895.3274915}
Jamyoung Koo, Junghoon Seo, Seunghyun Jeon, Jeongyeol Choe, and Taegyun Jeon.
\newblock Rbox-cnn: rotated bounding box based cnn for ship detection in remote sensing image.
\newblock In \emph{Proceedings of the 26th ACM SIGSPATIAL International Conference on Advances in Geographic Information Systems}, SIGSPATIAL '18, page 420–423, New York, NY, USA, 2018. Association for Computing Machinery.
\newblock ISBN 9781450358897.
\newblock \doi{10.1145/3274895.3274915}.

\bibitem[{Xu} et~al.(2023){Xu}, {Ding}, {Wang}, {Yang}, {Yu}, {Yu}, and {Xia}]{xu2023coarsetodynamic}
Chang {Xu}, Jian {Ding}, Jinwang {Wang}, Wen {Yang}, Huai {Yu}, Lei {Yu}, and Gui-Song {Xia}.
\newblock Dynamic coarse-to-fine learning for oriented tiny object detection.
\newblock In \emph{Proceedings of the IEEE/CVF Conference on Computer Vision and Pattern Recognition}, 2023.

\bibitem[Han et~al.(2021{\natexlab{a}})Han, Ding, Li, and Xia]{han2021align}
Jiaming Han, Jian Ding, Jie Li, and Gui-Song Xia.
\newblock Align deep features for oriented object detection.
\newblock \emph{IEEE Transactions on Geoscience and Remote Sensing}, 60:\penalty0 1--11, 2021{\natexlab{a}}.

\bibitem[Han et~al.(2021{\natexlab{b}})Han, Ding, Xue, and Xia]{han2021redet}
Jiaming Han, Jian Ding, Nan Xue, and Gui-Song Xia.
\newblock Redet: A rotation-equivariant detector for aerial object detection.
\newblock In \emph{Proceedings of the IEEE/CVF Conference on Computer Vision and Pattern Recognition}, pages 2786--2795, 2021{\natexlab{b}}.

\bibitem[Xie et~al.(2021)Xie, Cheng, Wang, Yao, and Han]{xie2021orientedrcnn}
Xingxing Xie, Gong Cheng, Jiabao Wang, Xiwen Yao, and Junwei Han.
\newblock Oriented r-cnn for object detection.
\newblock In \emph{Proceedings of the IEEE/CVF International Conference on Computer Vision}, pages 3520--3529, 2021.

\bibitem[Yang et~al.(2021{\natexlab{a}})Yang, Yang, Yang, Ming, Wang, Tian, and Yan]{yang2021kld}
Xue Yang, Xiaojiang Yang, Jirui Yang, Qi~Ming, Wentao Wang, Qi~Tian, and Junchi Yan.
\newblock Learning high-precision bounding box for rotated object detection via kullback-leibler divergence.
\newblock In M.~Ranzato, A.~Beygelzimer, Y.~Dauphin, P.S. Liang, and J.~Wortman Vaughan, editors, \emph{Advances in Neural Information Processing Systems}, volume~34, pages 18381--18394. Curran Associates, Inc., 2021{\natexlab{a}}.

\bibitem[Yang et~al.(2021{\natexlab{b}})Yang, Yan, Ming, Wang, Zhang, and Tian]{yang2021gwd}
Xue Yang, Junchi Yan, Qi~Ming, Wentao Wang, Xiaopeng Zhang, and Qi~Tian.
\newblock Rethinking rotated object detection with gaussian wasserstein distance loss.
\newblock In Marina Meila and Tong Zhang, editors, \emph{Proceedings of the 38th International Conference on Machine Learning}, volume 139 of \emph{Proceedings of Machine Learning Research}, pages 11830--11841. PMLR, 18--24 Jul 2021{\natexlab{b}}.

\bibitem[Yang et~al.(2023{\natexlab{b}})Yang, Zhou, Zhang, Yang, Wang, Yan, ZHANG, and Tian]{yang2023the}
Xue Yang, Yue Zhou, Gefan Zhang, Jirui Yang, Wentao Wang, Junchi Yan, XIAOPENG ZHANG, and Qi~Tian.
\newblock The {KFI}ou loss for rotated object detection.
\newblock In \emph{The Eleventh International Conference on Learning Representations}, 2023{\natexlab{b}}.

\bibitem[Yu and Da(2023)]{Yu_Da_2022}
Yi~Yu and Feipeng Da.
\newblock Phase-shifting coder: Predicting accurate orientation in oriented object detection.
\newblock In \emph{2023 IEEE/CVF Conference on Computer Vision and Pattern Recognition (CVPR)}, 2023.

\bibitem[Yang and Yan(2020)]{yang2020arbitrary}
Xue Yang and Junchi Yan.
\newblock Arbitrary-oriented object detection with circular smooth label.
\newblock In Andrea Vedaldi, Horst Bischof, Thomas Brox, and Jan-Michael Frahm, editors, \emph{Computer Vision -- ECCV 2020}, pages 677--694, Cham, 2020. Springer International Publishing.
\newblock ISBN 978-3-030-58598-3.

\bibitem[Yang et~al.(2021{\natexlab{c}})Yang, Hou, Zhou, Wang, and Yan]{yang2021dense}
Xue Yang, Liping Hou, Yue Zhou, Wentao Wang, and Junchi Yan.
\newblock Dense label encoding for boundary discontinuity free rotation detection.
\newblock In \emph{Proceedings of the IEEE/CVF conference on computer vision and pattern recognition}, pages 15819--15829, 2021{\natexlab{c}}.

\bibitem[Qian et~al.(2021)Qian, Yang, Peng, Yan, and Guo]{Qian_Yang_Peng_Yan_Guo_2021}
Wen Qian, Xue Yang, Silong Peng, Junchi Yan, and Yue Guo.
\newblock Learning modulated loss for rotated object detection.
\newblock \emph{Proceedings of the AAAI Conference on Artificial Intelligence}, 35\penalty0 (3):\penalty0 2458--2466, May 2021.
\newblock \doi{10.1609/aaai.v35i3.16347}.

\bibitem[Carion et~al.(2020)Carion, Massa, Synnaeve, Usunier, Kirillov, and Zagoruyko]{carion2020detr}
Nicolas Carion, Francisco Massa, Gabriel Synnaeve, Nicolas Usunier, Alexander Kirillov, and Sergey Zagoruyko.
\newblock End-to-end object detection with transformers.
\newblock In \emph{Computer Vision--ECCV 2020: 16th European Conference, Glasgow, UK, August 23--28, 2020, Proceedings, Part I 16}, pages 213--229. Springer, 2020.

\bibitem[Zhu et~al.(2021)Zhu, Su, Lu, Li, Wang, and Dai]{zhu2020deformable}
Xizhou Zhu, Weijie Su, Lewei Lu, Bin Li, Xiaogang Wang, and Jifeng Dai.
\newblock Deformable {\{}detr{\}}: Deformable transformers for end-to-end object detection.
\newblock In \emph{International Conference on Learning Representations}, 2021.

\bibitem[Liu et~al.(2022)Liu, Li, Zhang, Yang, Qi, Su, Zhu, and Zhang]{liu2022dab}
Shilong Liu, Feng Li, Hao Zhang, Xiao Yang, Xianbiao Qi, Hang Su, Jun Zhu, and Lei Zhang.
\newblock {DAB}-{DETR}: Dynamic anchor boxes are better queries for {DETR}.
\newblock In \emph{International Conference on Learning Representations}, 2022.

\bibitem[Zhang et~al.(2023)Zhang, Li, Liu, Zhang, Su, Zhu, Ni, and Shum]{zhang2022dino}
Hao Zhang, Feng Li, Shilong Liu, Lei Zhang, Hang Su, Jun Zhu, Lionel Ni, and Heung-Yeung Shum.
\newblock {DINO}: {DETR} with improved denoising anchor boxes for end-to-end object detection.
\newblock In \emph{The Eleventh International Conference on Learning Representations}, 2023.

\bibitem[Lin et~al.(2014)Lin, Maire, Belongie, Hays, Perona, Ramanan, Doll{\'a}r, and Zitnick]{lin2014coco}
Tsung-Yi Lin, Michael Maire, Serge Belongie, James Hays, Pietro Perona, Deva Ramanan, Piotr Doll{\'a}r, and C~Lawrence Zitnick.
\newblock Microsoft coco: Common objects in context.
\newblock In \emph{Computer Vision--ECCV 2014: 13th European Conference, Zurich, Switzerland, September 6-12, 2014, Proceedings, Part V 13}, pages 740--755. Springer, 2014.

\bibitem[Yang et~al.(2022)Yang, Li, Dai, and Gao]{yang2022focal}
Jianwei Yang, Chunyuan Li, Xiyang Dai, and Jianfeng Gao.
\newblock Focal modulation networks.
\newblock \emph{Advances in Neural Information Processing Systems}, 35:\penalty0 4203--4217, 2022.

\bibitem[Zong et~al.(2022)Zong, Song, and Liu]{zong2022codetr}
Zhuofan Zong, Guanglu Song, and Yu~Liu.
\newblock Detrs with collaborative hybrid assignments training.
\newblock \emph{arXiv preprint arXiv:2211.12860}, 2022.

\bibitem[Wang et~al.(2022{\natexlab{a}})Wang, Dai, Chen, Huang, Li, Zhu, Hu, Lu, Lu, Li, et~al.]{wang2022internimage}
Wenhai Wang, Jifeng Dai, Zhe Chen, Zhenhang Huang, Zhiqi Li, Xizhou Zhu, Xiaowei Hu, Tong Lu, Lewei Lu, Hongsheng Li, et~al.
\newblock Internimage: Exploring large-scale vision foundation models with deformable convolutions.
\newblock \emph{arXiv preprint arXiv:2211.05778}, 2022{\natexlab{a}}.

\bibitem[Ma et~al.(2021)Ma, Mao, Zheng, Gao, Wang, Han, Ding, Zhang, and Doermann]{ma2021o2d}
Teli Ma, Mingyuan Mao, Honghui Zheng, Peng Gao, Xiaodi Wang, Shumin Han, Errui Ding, Baochang Zhang, and David Doermann.
\newblock Oriented object detection with transformer.
\newblock \emph{arXiv preprint arXiv:2106.03146}, 2021.

\bibitem[Dai et~al.(2022)Dai, Liu, Tang, Wu, and Song]{dai2022ao2}
Linhui Dai, Hong Liu, Hao Tang, Zhiwei Wu, and Pinhao Song.
\newblock Ao2-detr: Arbitrary-oriented object detection transformer.
\newblock \emph{IEEE Transactions on Circuits and Systems for Video Technology}, 2022.

\bibitem[Zeng et~al.(2024)Zeng, Chen, Yang, Li, and Yan]{zeng2023ars}
Ying Zeng, Yushi Chen, Xue Yang, Qingyun Li, and Junchi Yan.
\newblock Ars-detr: Aspect ratio-sensitive detection transformer for aerial oriented object detection.
\newblock \emph{IEEE Transactions on Geoscience and Remote Sensing}, 62:\penalty0 1--15, 2024.
\newblock \doi{10.1109/TGRS.2024.3364713}.

\bibitem[Zhou et~al.(2023)Zhou, Yu, Wang, and Wang]{zhou2023d2q}
Qiang Zhou, Chaohui Yu, Zhibin Wang, and Fan Wang.
\newblock D2q-detr: Decoupling and dynamic queries for oriented object detection with transformers.
\newblock In \emph{IEEE International Conference on Acoustics, Speech and Signal Processing (ICASSP)}, 2023.

\bibitem[Yang et~al.(2021{\natexlab{d}})Yang, Hou, Zhou, Wang, and Yan]{9578392}
Xue Yang, Liping Hou, Yue Zhou, Wentao Wang, and Junchi Yan.
\newblock Dense label encoding for boundary discontinuity free rotation detection.
\newblock In \emph{2021 IEEE/CVF Conference on Computer Vision and Pattern Recognition (CVPR)}, pages 15814--15824, 2021{\natexlab{d}}.
\newblock \doi{10.1109/CVPR46437.2021.01556}.

\bibitem[Cai et~al.(2024)Cai, Lai, Wang, Wang, Sun, and Yao]{cai2024poly}
Xinhao Cai, Qiuxia Lai, Yuwei Wang, Wenguan Wang, Zeren Sun, and Yazhou Yao.
\newblock Poly kernel inception network for remote sensing detection.
\newblock In \emph{Proceedings of the IEEE conference on computer vision and pattern recognition}, 2024.

\bibitem[Xu et~al.(2024)Xu, Liu, Xu, Ma, Zhu, Yan, and Dai]{xu2023rethinking}
Hang Xu, Xinyuan Liu, Haonan Xu, Yike Ma, Zunjie Zhu, Chenggang Yan, and Feng Dai.
\newblock Rethinking boundary discontinuity problem for oriented object detection.
\newblock In \emph{Proceedings of the IEEE conference on computer vision and pattern recognition}, 2024.

\bibitem[Xiao et~al.(2024)Xiao, Yang, Yang, Mu, Yan, and Hu]{xiao2024theoretically}
Zikai Xiao, Guo-Ye Yang, Xue Yang, Tai-Jiang Mu, Junchi Yan, and Shi-min Hu.
\newblock Theoretically achieving continuous representation of oriented bounding boxes.
\newblock In \emph{Proceedings of the IEEE conference on computer vision and pattern recognition}, 2024.

\bibitem[Yang et~al.(2023{\natexlab{c}})Yang, Zhou, Zhang, Yang, Wang, Yan, Zhang, and Tian]{yang2022kfiou}
Xue Yang, Yue Zhou, Gefan Zhang, Jitui Yang, Wentao Wang, Junchi Yan, Xiaopeng Zhang, and Qi~Tian.
\newblock The {KFI}ou loss for rotated object detection.
\newblock In \emph{The Eleventh International Conference on Learning Representations}, 2023{\natexlab{c}}.

\bibitem[Zeng et~al.(2023)Zeng, Ran, Zhu, Gao, Qiu, and Chen]{10214368dynamiccascade}
Qiaolin Zeng, Xiang Ran, Hao Zhu, Yanghua Gao, Xinfa Qiu, and Liangfu Chen.
\newblock Dynamic cascade query selection for oriented object detection.
\newblock \emph{IEEE Geoscience and Remote Sensing Letters}, 20:\penalty0 1--5, 2023.
\newblock \doi{10.1109/LGRS.2023.3304023}.

\bibitem[Feng and Wang(2024)]{10387417psdsq}
Shiyang Feng and Bin Wang.
\newblock Psd-sq: Point set decoding based on semantic query for object detection in remote sensing images.
\newblock \emph{IEEE Transactions on Geoscience and Remote Sensing}, 62:\penalty0 1--12, 2024.
\newblock \doi{10.1109/TGRS.2024.3352011}.

\bibitem[Ma et~al.(2024)Ma, Lv, and Zhong]{10474035qetr}
Xinyu Ma, Pengyuan Lv, and Yanfei Zhong.
\newblock Qetr: A query-enhanced transformer for remote sensing image object detection.
\newblock \emph{IEEE Geoscience and Remote Sensing Letters}, 21:\penalty0 1--5, 2024.
\newblock \doi{10.1109/LGRS.2024.3378531}.

\bibitem[Hu et~al.(2023)Hu, Gao, Zhang, Wang, Wang, Yang, Li, and Li]{hu2023emo2}
Zibo Hu, Kun Gao, Xiaodian Zhang, Junwei Wang, Hong Wang, Zhijia Yang, Chenrui Li, and Wei Li.
\newblock Emo2-detr: Efficient-matching oriented object detection with transformers.
\newblock \emph{IEEE Transactions on Geoscience and Remote Sensing}, 2023.

\bibitem[Li et~al.(2022)Li, Zhang, Liu, Guo, Ni, and Zhang]{li2022dn}
Feng Li, Hao Zhang, Shilong Liu, Jian Guo, Lionel~M Ni, and Lei Zhang.
\newblock Dn-detr: Accelerate detr training by introducing query denoising.
\newblock In \emph{Proceedings of the IEEE/CVF Conference on Computer Vision and Pattern Recognition}, pages 13619--13627, 2022.

\bibitem[Lin et~al.(2017)Lin, Goyal, Girshick, He, and Doll{\'a}r]{lin2017focal}
Tsung-Yi Lin, Priya Goyal, Ross Girshick, Kaiming He, and Piotr Doll{\'a}r.
\newblock Focal loss for dense object detection.
\newblock In \emph{Proceedings of the IEEE international conference on computer vision}, pages 2980--2988, 2017.

\bibitem[Rezatofighi et~al.(2019)Rezatofighi, Tsoi, Gwak, Sadeghian, Reid, and Savarese]{rezatofighi2019giou}
Hamid Rezatofighi, Nathan Tsoi, JunYoung Gwak, Amir Sadeghian, Ian Reid, and Silvio Savarese.
\newblock Generalized intersection over union: A metric and a loss for bounding box regression.
\newblock In \emph{Proceedings of the IEEE/CVF conference on computer vision and pattern recognition}, pages 658--666, 2019.

\bibitem[Attouch et~al.(1991)Attouch, Lucchetti, and Wets]{attouch1991topology}
Hedy Attouch, Roberto Lucchetti, and Roger J-B Wets.
\newblock The topology of the $\rho$-hausdorff distance.
\newblock \emph{Annali di Matematica pura ed applicata}, 160\penalty0 (1):\penalty0 303--320, 1991.

\bibitem[Cha et~al.(2024)Cha, Seo, and Lee]{cha2023billion}
Keumgang Cha, Junghoon Seo, and Taekyung Lee.
\newblock A billion-scale foundation model for remote sensing images.
\newblock \emph{IEEE Journal of Selected Topics in Applied Earth Observations and Remote Sensing}, pages 1--17, 2024.
\newblock \doi{10.1109/JSTARS.2024.3401772}.

\bibitem[He et~al.(2017)He, Gkioxari, Doll{\'a}r, and Girshick]{he2017mask}
Kaiming He, Georgia Gkioxari, Piotr Doll{\'a}r, and Ross Girshick.
\newblock Mask r-cnn.
\newblock In \emph{Proceedings of the IEEE international conference on computer vision}, pages 2961--2969, 2017.

\bibitem[Chen et~al.(2019)Chen, Pang, Wang, Xiong, Li, Sun, Feng, Liu, Shi, Ouyang, Loy, and Lin]{chen2019hybrid}
Kai Chen, Jiangmiao Pang, Jiaqi Wang, Yu~Xiong, Xiaoxiao Li, Shuyang Sun, Wansen Feng, Ziwei Liu, Jianping Shi, Wanli Ouyang, Chen~Change Loy, and Dahua Lin.
\newblock Hybrid task cascade for instance segmentation.
\newblock In \emph{Proceedings of the IEEE/CVF conference on computer vision and pattern recognition}, pages 4974--4983, 2019.

\bibitem[Lyu et~al.(2022)Lyu, Zhang, Huang, Zhou, Wang, Liu, Zhang, and Chen]{lyu2022rtmdet}
Chengqi Lyu, Wenwei Zhang, Haian Huang, Yue Zhou, Yudong Wang, Yanyi Liu, Shilong Zhang, and Kai Chen.
\newblock Rtmdet: An empirical study of designing real-time object detectors.
\newblock \emph{arXiv preprint arXiv:2212.07784}, 2022.

\bibitem[Wang et~al.(2022{\natexlab{b}})Wang, Zhang, Xu, Zhang, Du, Tao, and Zhang]{wang2022vitaers}
Di~Wang, Qiming Zhang, Yufei Xu, Jing Zhang, Bo~Du, Dacheng Tao, and Liangpei Zhang.
\newblock Advancing plain vision transformer towards remote sensing foundation model.
\newblock \emph{IEEE Transactions on Geoscience and Remote Sensing}, 2022{\natexlab{b}}.

\bibitem[Cheng et~al.(2022)Cheng, Wang, Li, Xie, Lang, Yao, and Han]{cheng2022diorr}
Gong Cheng, Jiabao Wang, Ke~Li, Xingxing Xie, Chunbo Lang, Yanqing Yao, and Junwei Han.
\newblock Anchor-free oriented proposal generator for object detection.
\newblock \emph{IEEE Transactions on Geoscience and Remote Sensing}, 60:\penalty0 1--11, 2022.

\bibitem[Dosovitskiy et~al.(2020)Dosovitskiy, Beyer, Kolesnikov, Weissenborn, Zhai, Unterthiner, Dehghani, Minderer, Heigold, Gelly, et~al.]{dosovitskiy2020vit}
Alexey Dosovitskiy, Lucas Beyer, Alexander Kolesnikov, Dirk Weissenborn, Xiaohua Zhai, Thomas Unterthiner, Mostafa Dehghani, Matthias Minderer, Georg Heigold, Sylvain Gelly, et~al.
\newblock An image is worth 16x16 words: Transformers for image recognition at scale.
\newblock \emph{arXiv preprint arXiv:2010.11929}, 2020.

\bibitem[Ren et~al.(2015)Ren, He, Girshick, and Sun]{ren2015faster}
Shaoqing Ren, Kaiming He, Ross Girshick, and Jian Sun.
\newblock Faster r-cnn: Towards real-time object detection with region proposal networks.
\newblock \emph{Advances in neural information processing systems}, 28, 2015.

\bibitem[Zhou et~al.(2019)Zhou, Fang, Song, Guan, Yin, Dai, and Yang]{zhou2019iou}
Dingfu Zhou, Jin Fang, Xibin Song, Chenye Guan, Junbo Yin, Yuchao Dai, and Ruigang Yang.
\newblock Iou loss for 2d/3d object detection.
\newblock In \emph{2019 international conference on 3D vision (3DV)}, pages 85--94. IEEE, 2019.

\bibitem[Zhou et~al.(2022)Zhou, Yang, Zhang, Wang, Liu, Hou, Jiang, Liu, Yan, Lyu, Zhang, and Chen]{zhou2022mmrotate}
Yue Zhou, Xue Yang, Gefan Zhang, Jiabao Wang, Yanyi Liu, Liping Hou, Xue Jiang, Xingzhao Liu, Junchi Yan, Chengqi Lyu, Wenwei Zhang, and Kai Chen.
\newblock Mmrotate: A rotated object detection benchmark using pytorch.
\newblock In \emph{Proceedings of the 30th ACM International Conference on Multimedia}, pages 7331--7334, 2022.

\bibitem[He et~al.(2016)He, Zhang, Ren, and Sun]{he2016resnet}
Kaiming He, Xiangyu Zhang, Shaoqing Ren, and Jian Sun.
\newblock Deep residual learning for image recognition.
\newblock In \emph{Proceedings of the IEEE conference on computer vision and pattern recognition}, pages 770--778, 2016.

\bibitem[Liu et~al.(2021)Liu, Lin, Cao, Hu, Wei, Zhang, Lin, and Guo]{liu2021swin}
Ze~Liu, Yutong Lin, Yue Cao, Han Hu, Yixuan Wei, Zheng Zhang, Stephen Lin, and Baining Guo.
\newblock Swin transformer: Hierarchical vision transformer using shifted windows.
\newblock In \emph{Proceedings of the IEEE/CVF international conference on computer vision}, pages 10012--10022, 2021.

\bibitem[Deng et~al.(2009)Deng, Dong, Socher, Li, Li, and Fei-Fei]{deng2009imagenet}
Jia Deng, Wei Dong, Richard Socher, Li-Jia Li, Kai Li, and Li~Fei-Fei.
\newblock Imagenet: A large-scale hierarchical image database.
\newblock In \emph{2009 IEEE conference on computer vision and pattern recognition}, pages 248--255. Ieee, 2009.

\bibitem[Kingma and Ba(2014)]{kingma2014adam}
Diederik~P Kingma and Jimmy Ba.
\newblock Adam: A method for stochastic optimization.
\newblock \emph{arXiv preprint arXiv:1412.6980}, 2014.

\bibitem[Loshchilov and Hutter(2017)]{loshchilov2017adamw}
Ilya Loshchilov and Frank Hutter.
\newblock Decoupled weight decay regularization.
\newblock \emph{arXiv preprint arXiv:1711.05101}, 2017.

\bibitem[Yang et~al.(2021{\natexlab{e}})Yang, Yan, Feng, and He]{yang2021r3det}
Xue Yang, Junchi Yan, Ziming Feng, and Tao He.
\newblock R3det: Refined single-stage detector with feature refinement for rotating object.
\newblock In \emph{Proceedings of the AAAI conference on artificial intelligence}, volume~35, pages 3163--3171, 2021{\natexlab{e}}.

\end{thebibliography}

}
\end{document}